\documentclass[10pt,twocolumn,letterpaper]{article}

\PassOptionsToPackage{table}{xcolor}
\usepackage[pagenumbers]{cvpr}
\usepackage[T1]{fontenc}
\usepackage{amsmath,amssymb,booktabs,graphicx,multirow}
\usepackage{tikz}
\usetikzlibrary{arrows.meta,positioning,calc,fit}
\usepackage{enumitem}

\definecolor{traceblue}{HTML}{2166AC}
\definecolor{traceorange}{HTML}{B35806}
\definecolor{tracegreen}{HTML}{1B7837}

\usepackage{booktabs}
\usepackage{multirow}
\usepackage{tabularx}
\providecolor{oursgray}{gray}{0.94}

\makeatletter
\newsavebox{\ablationstrutbox}
\newcommand{\ablationwhiterow}{%
  \noalign{%
    \global\setbox\ablationstrutbox=\copy\@arstrutbox
    \global\setbox\@arstrutbox=\hbox{%
      \vrule height 2pt depth 0pt width 0pt}%
  }%
  \rowcolor{white}
  & & & & & & & & \\
  \noalign{%
    \global\setbox\@arstrutbox=\copy\ablationstrutbox
  }%
}
\makeatother

\usepackage{booktabs}
\usepackage{makecell}

\usepackage{algorithm}
\usepackage{algpseudocode}
\usepackage[pagebackref,breaklinks,colorlinks,allcolors=traceblue]{hyperref}

\title{Look Back Before You Change: Exploring Multimodal LLM Feedback for Training-free Open-Vocabulary Change Detection}
\author{Qirui Guo\textsuperscript{1}, Libo Wang\textsuperscript{2,1},
Yuanchao Su\textsuperscript{3}, Xianping Ma\textsuperscript{4},
and Xiaokang Zhang\textsuperscript{5}\\
\textsuperscript{1}Nanjing University of Information Science and Technology\\
\textsuperscript{2}University of Macau \quad
\textsuperscript{3}Xi'an University of Science and Technology\\
\textsuperscript{4}Southwest Jiaotong University \quad
\textsuperscript{5}Wuhan University}
\hypersetup{
    pdftitle={Look Back Before You Change: Exploring Multimodal LLM Feedback for Training-free Open-Vocabulary Change Detection},
    pdfauthor={Qirui Guo, Libo Wang, Yuanchao Su, Xianping Ma, Xiaokang Zhang}
}

\begin{document}
\maketitle
\begin{abstract}
Open-vocabulary change detection (OVCD) localizes changes specified by natural language queries in bi-temporal remote sensing imagery. However, visual similarities between land-cover categories and cross-temporal appearance variations can cause foundation models to misjudge target category or mask coverage, leading to false positives or false negatives. We propose LookBackOVCD, a training-free OVCD framework that uses feedback from a multimodal large language model (MLLM) to correct initial predictions before temporal comparison. The MLLM reviews both original images and relevant local regions to reassess target predictions by examining object attributes, surrounding context, and mask coverage. The resulting feedback guides local corrections to semantic responses and prompts a segmentation model to generate or refine masks in each image. A shared evidence pool stores and reuses observations for reference in subsequent judgments. The updated predictions are then compared across dates to produce the change map. Experiments on LEVIR-CD, WHU-CD, SECOND, and SC-SCD demonstrate that LookBackOVCD outperforms existing training-free methods in both building and multi-class change detection.
\end{abstract}
\section{Introduction}
\label{sec:introduction}

Open-vocabulary change detection (OVCD) localizes changes specified by natural language queries in bi-temporal remote sensing imagery. Training-free OVCD pipelines combine pretrained foundation models for mask generation, category identification, and temporal comparison \cite{li2026dynamicearth}. However, foundation models still have limitations in recognizing targets in remote sensing imagery. Similar appearances across land-cover types and cross-temporal appearance variations of the same object can both make recognition more difficult. Errors in category identification or mask coverage in either image can propagate into temporal comparison, resulting in false positives or false negatives.

\begin{figure}[t]
    \centering
    \includegraphics[width=\linewidth]{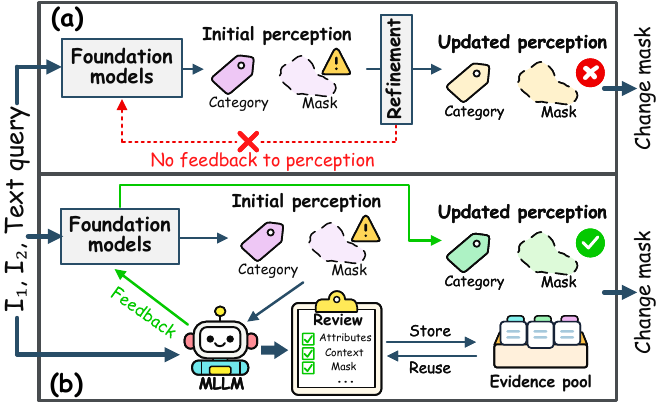}
    \caption{MLLM feedback in open-vocabulary change detection.
    (a) An initial recognition error can persist after refinement and affect the change mask.
    (b) Our framework uses an MLLM to reassess target category and mask coverage in both images and guide corrections before temporal comparison. A shared evidence pool retains previous observations for reference in subsequent judgments.}
    \label{fig:teaser}
\end{figure}

Recent OVCD methods address these challenges from different perspectives. OpenDPR constructs category-specific visual prototypes from diffusion-generated images and retrieves matching prototypes for candidate change regions, improving recognition through comparison in the visual feature space \cite{Guo_2026_CVPR}. CoRegOVCD combines semantic calibration with structural verification to refine change predictions \cite{tang2026coregovcd}, while ReA-OVCD assesses the semantic and spatial reliability of candidate changes \cite{zhu2026rea}. However, initial recognition errors may still persist through subsequent refinement and affect the final change mask (Fig.~\ref{fig:teaser}(a)).

Visual reasoning studies offer inspiration for addressing these errors by checking existing judgments against the original image. Visual search methods revisit relevant regions to obtain details needed for the current task \cite{wu2024vstar,shen2025zoomeye}. The strategy in VISCO checks existing responses against the original image and uses the identified inconsistencies to guide correction \cite{wu2025visco}. ECRD examines the image to guide the decisions of the model when existing evidence is insufficient \cite{zhang2026see}. In OVCD, revisiting the original images allows initial predictions to be checked against the appearance of targets, surrounding context, and visible boundaries. This review can identify category and coverage errors that persist in the predictions. We therefore review both original images with a multimodal large language model (MLLM) and use its feedback to correct semantic responses and masks before temporal comparison (Fig.~\ref{fig:teaser}(b)). These MLLMs can interpret text queries together with visual information and reason about objects in their surrounding context. Given initial predictions, an MLLM can assess category membership and examine how predicted regions correspond to visible targets. These judgments indicate which objects should be retained or removed, where coverage requires adjustment, and which targets need to be recovered. We therefore combine MLLM review with promptable segmentation: the MLLM identifies the required corrections and provides spatial guidance, while a dedicated segmentation model generates or refines the corresponding masks. This cooperation translates visual judgments into pixel-level updates.

Building on these considerations, we propose LookBackOVCD, a training-free OVCD framework that uses MLLM feedback to refine target predictions before temporal comparison. SAM~3 generates semantic responses and segmentation masks for the queried concept in each image. Qwen2.5-VL jointly examines both full images to assess initial predictions. It then reviews regions requiring verification locally, avoiding repeated inspection of regions already resolved globally. Local review combines full-image information with paired local crops to reassess target category and mask coverage through object attributes and surrounding context, and to identify potentially overlooked targets. Observations are stored in a shared evidence pool for subsequent judgments. The resulting feedback guides local updates to the semantic responses and prompts SAM~3 to generate or refine masks in the corresponding image. We compare the updated predictions across dates to produce the change map for the queried concept.

Our main contributions are as follows.
\begin{itemize}
    \item We introduce a MLLM-based visual verification mechanism into OVCD before temporal comparison. By reasoning about object attributes and scene context in both original images, the MLLM reassesses the category membership and mask coverage of initial target predictions.

    \item We propose LookBackOVCD, a training-free framework that uses Qwen2.5-VL feedback from global and local review to guide local corrections of SAM~3 predictions. A shared evidence pool retains observations from previous reviews for reference in subsequent judgments.

    \item Experiments on multiple datasets demonstrate the effectiveness of our method for both building and multi-class change detection.
\end{itemize}
\section{Related Work}
\label{sec:relatedwork}

\subsection{Open-Vocabulary Change Detection}
DynamicEarth combines frozen foundation models in training-free M-C-I and I-M-C pipelines, which differ in the order of category identification and temporal comparison \cite{li2026dynamicearth}. OmniOVCD uses SAM~3 alone for instance-level change comparison \cite{zhang2026omniovcd}, while AdaptOVCD combines models through adaptive information fusion \cite{dou2026adaptovcd}. OpenDPR improves category recognition by matching change proposals with diffusion-generated visual prototypes \cite{Guo_2026_CVPR}. CoRegOVCD combines posterior calibration with structural verification \cite{tang2026coregovcd}, while ReA-OVCD assesses the semantic and spatial reliability of candidate changes \cite{zhu2026rea}. MemOVCD uses bidirectional memory propagation and global--local prediction fusion to improve consistency \cite{kuang2026memovcd}. Despite these advances, foundation models can still misidentify queried targets or produce masks with incorrect coverage. Our framework uses an MLLM to reassess these predictions against visual evidence in both original images and provide feedback for correction before temporal comparison.

\subsection{Foundation Models}
Foundation models support OVCD through mask generation, visual feature extraction, and text-driven recognition. SAM generates masks from spatial prompts \cite{kirillov2023segment}, while SAM~2 extends promptable segmentation to videos through streaming memory \cite{ravi2025sam}. SAM~3 supports text and exemplar prompts to detect and segment queried concepts \cite{carion2026sam}. The DINO family learns transferable dense features through self-supervision \cite{caron2021emerging,oquab2024dinov2,simeoni2026dinov3}. CLIP aligns image and text representations for zero-shot recognition \cite{radford2021learning}. SegEarth-OV improves CLIP-based remote sensing segmentation through feature upsampling and global bias alleviation \cite{li2025segearthov}, while SegEarth-OV3 adapts SAM~3 through mask fusion and presence-guided filtering \cite{li2025segearthov3}. These models provide complementary representations, including object masks, dense visual features, and semantic responses, for recognizing queried targets and comparing predictions across dates.

\subsection{Multimodal Large Language Models}
BLIP-2 connects frozen image encoders and language models through a querying transformer \cite{li2023blip2}, while LLaVA improves instruction following through visual instruction tuning \cite{liu2023visual}. Shikra supports coordinate-based dialogue \cite{chen2023shikra}, and Ferret combines coordinates with visual features for regional interaction and grounding \cite{you2024ferret}. Qwen2.5-VL supports fine-grained recognition and object localization using boxes and points \cite{bai2025qwen25vl}. LISA generates segmentation masks from complex, implicit queries \cite{lai2024lisa}, while GLaMM associates generated text with object masks \cite{rasheed2024glamm}. Recent studies also examine visual evidence gathering and verification. ZoomEye searches image regions for relevant details through a tree search \cite{shen2025zoomeye}, while the LookBack strategy in VISCO revisits images to verify visual claims and guide subsequent correction \cite{wu2025visco}. These capabilities support using MLLMs to reassess visual predictions and provide semantic and spatial feedback for refinement.
\section{Method}
\label{sec:method}

\begin{figure*}[t]
    \centering
    \includegraphics[width=\textwidth]{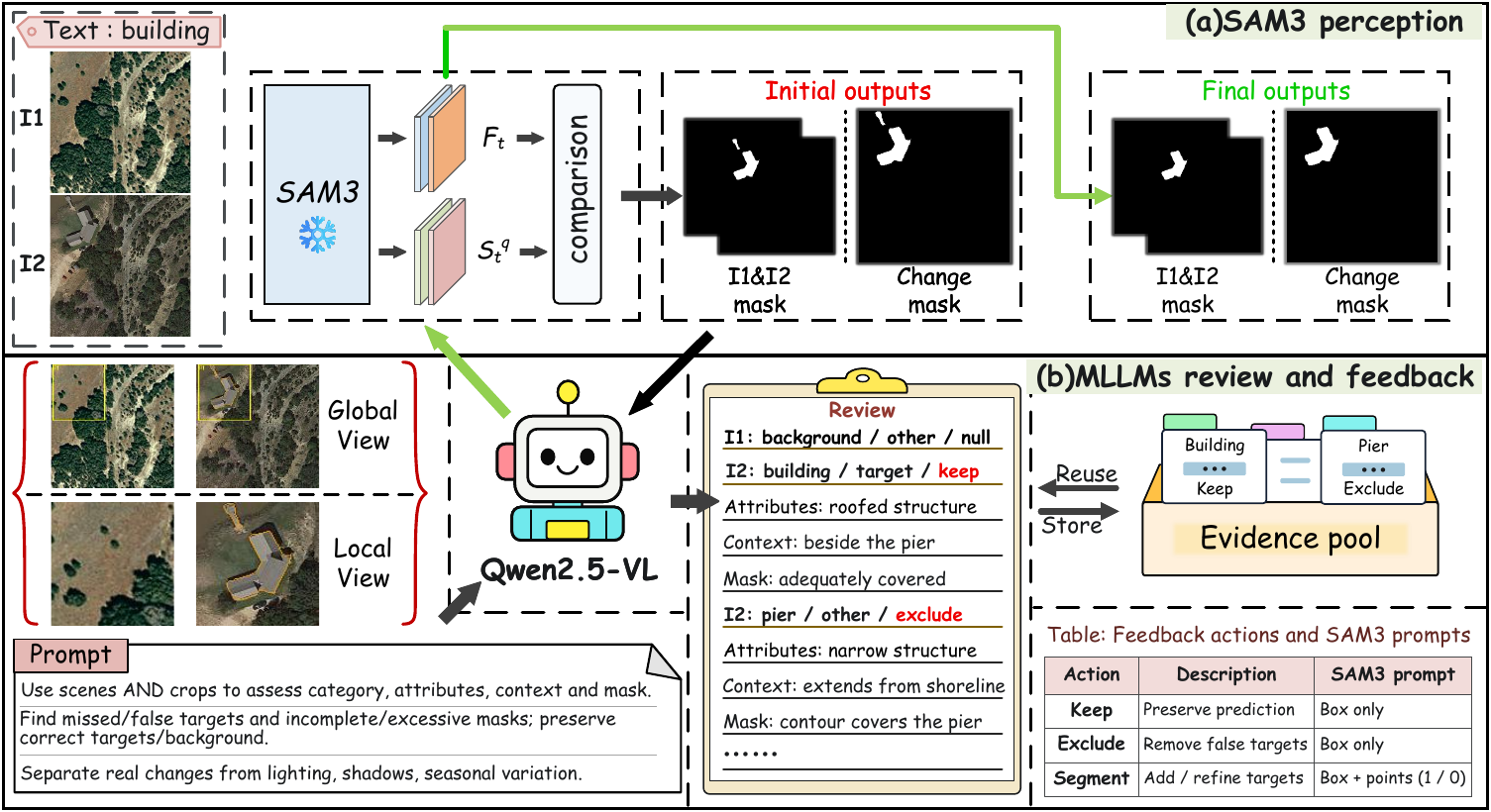}
    \caption{Overview of the proposed framework.}
    \label{fig:framework}
\end{figure*}

As illustrated in Fig.~\ref{fig:framework}, LookBackOVCD uses MLLM feedback to revise SAM~3 target predictions before temporal comparison. Given a text query, Qwen2.5-VL examines predictions from both images for category, mask, and omission errors. Its feedback guides corrections, while a shared evidence pool retains observations for subsequent judgments within the same image pair. The updated predictions are used for cross-temporal comparison to obtain the change map. All foundation models remain frozen.

\subsection{Initial Perception}

Given a bi-temporal image pair $I_1,I_2$ and a text query $q$, frozen SAM~3 extracts visual feature maps $F_t$ and produces query-conditioned instance predictions $S_t^q$ for each date $t\in\{1,2\}$. We represent the retained predictions as
$S_t^q=\{(p_{t,i},c_{t,i})\}_{i=1}^{N_t}$,
where $p_{t,i}$ is an instance mask probability map and $c_{t,i}$ is its confidence score. Their confidence-weighted probabilities are aggregated through a pixelwise maximum:
\begin{equation}
R_t^{(0)}(x)
=
\max_{1\leq i\leq N_t}
c_{t,i}\,p_{t,i}(x),
\label{eq:initial_response}
\end{equation}
where $x$ denotes a pixel and $R_t^{(0)}=0$ if no instance is retained. These continuous responses are updated through subsequent corrections.

Feature maps $F_t$ are reused to recompute object representations over updated masks in both images for subsequent temporal comparison.

\subsection{Multimodal Review and Feedback}

\noindent\textbf{Global-to-local verification.}
Let $R_t^{(k)}$ denote the target response, where $k=0$ corresponds to initial perception, and $k=1,2$ correspond to the outputs of global and local feedback, respectively. Its displayed mask is $M_t^{(k)}=\mathbf{1}[R_t^{(k)}>0.5]$. Connected regions in the union of the initial masks provide observation candidates. Spatially adjacent candidates are grouped using a minimum spanning tree over their bounding-box gaps, with the gap distribution determining the separation between groups. A group containing multiple candidates is divided spatially if it exceeds half the image width or height, while a single large candidate is retained intact. Context is added according to the member sizes within the applicable scale constraint. Each existing object is assigned to one review group, although neighboring crops may share contextual content.

Qwen first examines the full images, initial contours, and candidate regions to identify areas requiring closer inspection. It can propose regions containing potentially missed targets, extending local review beyond initial detections.

A candidate region bypasses local review when both dates are explicitly resolved. For each date, the outlined objects must match the query with adequate coverage and no suspected omission, or the region must contain neither a queried target nor a predicted contour. Ambiguous categories, unclear coverage, and unresolved omissions require local inspection. Resolved global judgments proceed directly to feedback execution. Temporal comparison follows global feedback or, when further review is needed, the feedback from the selected local regions.

\noindent\textbf{Local review.}
For window $w_r$, paired RGB crops with current mask boundaries form local views $V_1^{(r)}$ and $V_2^{(r)}$. The visual input is
$\mathcal{V}^{(r)}=(I_1,I_2,V_1^{(r)},V_2^{(r)})$.
The full images provide scene context, while the paired crops expose local appearance and response coverage. Object coordinates are provided as text references in the displayed crop coordinate system, linking each decision to its assigned object.

Shape, texture, and structural appearance support category identification, while surrounding land cover and relationships with nearby structures provide context. Comparing visible targets with contours reveals missing or excessive coverage. Each date is assessed separately, while paired views help distinguish structural differences from lighting or shadow variations.

Given the query, review instructions $\mathcal{T}$, and relevant historical evidence $\mathcal{E}^{(r)}$, Qwen2.5-VL, denoted by $\mathcal{Q}$, produces dated observations $\mathcal{O}^{(r)}$ and decisions $\mathcal{D}^{(r)}$:
\begin{equation}
\bigl(\mathcal{O}^{(r)},\mathcal{D}^{(r)}\bigr)
=
\mathcal{Q}\bigl(
\mathcal{V}^{(r)},q,\mathcal{T},\mathcal{E}^{(r)}
\bigr).
\label{eq:object_review}
\end{equation}
The model first generates a brief visual observation, then uses the same visual context to produce structured feedback. We write
$\mathcal{D}^{(r)}=\{d_j\}_{j=1}^{N_r}$,
where each decision specifies its date $t_j$, query membership, action $a_j$, and any required spatial prompts.

Each existing object assigned to the review region receives a separate decision, allowing different object types to be treated differently. An additional search identifies queried targets without contours and provides points inside them for SAM~3 segmentation. Valid object corrections are executed independently, while uncertain decisions are recorded as unresolved. The object decisions and missed-target search reuse the cached visual context without additional image encoding.

\noindent\textbf{Prediction correction.}
Query membership is classified as target, other, or uncertain. Resolved judgments lead to \textit{Keep}, \textit{Exclude}, or \textit{Segment} actions. \textit{Keep} confirms an existing target prediction, \textit{Exclude} removes responses associated with an identified non-target, and \textit{Segment} corrects incomplete or excessive target coverage. The MLLM determines what requires correction, while SAM~3 provides the corresponding pixel-level estimate.

For existing objects, feedback uses an object-specific box $b_j$ and, when needed, signed points
$\Pi_j=\{(\mathbf{p}_{jn},y_{jn})\}$,
where $y_{jn}\in\{0,1\}$. Positive points indicate target foreground to retain or recover, and negative points indicate incorrectly covered non-target pixels.

After mapping spatial prompts to original image coordinates, frozen SAM~3 produces a foreground probability map:
\begin{equation}
P_j
=
\sigma\bigl(
\mathcal{S}(I_{t_j};b_j,\Pi_j)
\bigr),
\label{eq:prompted_segmentation}
\end{equation}
where $\mathcal{S}$ denotes interactive segmentation logits and $\sigma$ is the sigmoid function. Point-only discovery uses $b_j=\varnothing$. \textit{Exclude} and confirmed-target \textit{Keep} use object boxes without point prompts.

For a potentially missed target, Qwen provides an interior positive point $\mathbf{p}_j$, and SAM~3 returns $L_j$ candidate probability maps $P_{j,\ell}$ with quality scores $s_{j,\ell}$. We select among masks containing the point and define the newly recovered foreground as
\begin{equation}
\begin{aligned}
\mathcal{C}_j
&=\{\ell\mid P_{j,\ell}(\mathbf{p}_j)>0.5\},\\
\ell_j^*
&=\arg\max_{\ell\in\mathcal{C}_j}s_{j,\ell},\\
\Omega_j^{\mathrm{miss}}
&=\{x\mid P_j(x)>0.5,\ R_{t_j}^{(k)}(x)\leq0.5\},
\end{aligned}
\label{eq:missed_target_recovery}
\end{equation}
where $P_j=P_{j,\ell_j^*}$. The \textit{Segment} action updates $R_{t_j}^{(k)}$ over $\Omega_j^{\mathrm{miss}}$ using $P_j$.

For \textit{Keep}, a box-prompted mask supplies a second boundary estimate. If its overlap with the current mask exceeds an IoU of $0.5$, the two responses are averaged along the current contour's immediate boundary. Other pixels remain unchanged.

Let $\Omega_j$ denote the editable support of a resolved decision. It is restricted to the object box for existing-object correction and to $\Omega_j^{\mathrm{miss}}$ for missed targets. For \textit{Exclude}, it covers the union of the old response mask and the prompted foreground within the object box. For \textit{Keep}, it contains only the boundary pixels eligible for fusion. The local proposal for $x\in\Omega_j$ is
\begin{equation}
\widetilde{R}_j(x)=
\begin{cases}
P_j(x), & a_j=\mathrm{Segment},\\
0, & a_j=\mathrm{Exclude},\\
\frac{1}{2}\bigl(R_{t_j}^{(k)}(x)+P_j(x)\bigr),
& a_j=\mathrm{Keep}.
\end{cases}
\label{eq:local_response_update}
\end{equation}
Responses outside $\Omega_j$ are preserved. Proposals are formed from the same current predictions, and compatible updates are merged into $R_t^{(k+1)}$. Conflicting pixels retain their previous values. Uncertain judgments and invalid spatial prompts are recorded as unresolved.

\noindent\textbf{Temporal comparison.}
Feedback revises target responses and, when masks change, their spatial support. For an object mask with weights $w_j$ at the feature resolution, paired representations are computed from $F_1$ and $F_2$ over the same support:
\begin{equation}
\begin{aligned}
\mathbf{z}_{t,j}
&=
\frac{\sum_u w_j(u)F_t(u)}{\sum_u w_j(u)},
\qquad t\in\{1,2\},\\
g_j
&=
\frac{1}{\pi}\arccos
\left(
\frac{\mathbf{z}_{1,j}^{\top}\mathbf{z}_{2,j}}
{\|\mathbf{z}_{1,j}\|_2\|\mathbf{z}_{2,j}\|_2}
\right).
\end{aligned}
\label{eq:updated_temporal_difference}
\end{equation}
Here, $w_j$ is obtained by resizing the object mask through area averaging, and $g_j$ measures its temporal feature difference. Initial instances establish comparison fields $G_t$, associating each pixel with the temporal difference of the instance contributing its strongest response. These fields are updated within accepted correction regions when object support changes. Pixelwise feature differences are used where valid object representations are unavailable.

After feedback, let $\widehat{R}_t$ and $\widehat{G}_t$ denote the resulting response and temporal comparison fields. The final prediction is
\begin{equation}
\begin{aligned}
t^*(x)
&=\arg\max_{t\in\{1,2\}}\widehat{R}_t(x),\\
D_q(x)
&=
\left|\widehat{R}_1(x)-\widehat{R}_2(x)\right|
\widehat{G}_{t^*(x)}(x),\\
\widehat{Y}_q(x)
&=
\mathbf{1}\left[
\left\lfloor255D_q(x)\right\rfloor>\tau
\right].
\end{aligned}
\label{eq:corrected_change_prediction}
\end{equation}
The response difference is therefore weighted by the temporal feature difference associated with the stronger response. Threshold $\tau$ converts the score into the change mask for the queried concept.

\subsection{Evidence Pool and Reuse}

We maintain a shared evidence pool $\mathcal{B}$ within each image pair and query. Global and local observations are stored with their dates, bounding boxes, and confirmed or uncertain review states. The pool is updated as review proceeds, making earlier observations available to subsequent regions.

Scene observations provide global context. Regional evidence is retrieved by spatial overlap with the current crop, using records matched to the current prediction state. Object references are restricted to the current crop. This retrieval preserves the spatial correspondence between observations and predictions. The selected evidence $\mathcal{E}^{(r)}$ in Eq.~\ref{eq:object_review} provides earlier observations of object characteristics, contextual relationships, and mask coverage. Qwen combines this evidence with the current visual input to assess target predictions and generate feedback.
\section{Experiments}
\label{sec:experiments}
\subsection{Datasets}
We evaluate our method on two building change detection datasets, LEVIR-CD and WHU-CD, and two semantic change detection datasets, SECOND and SC-SCD. LEVIR-CD \cite{chen2020spatial} contains 637 bi-temporal image pairs of $1024\times1024$ pixels at a spatial resolution of 0.5 m. WHU-CD \cite{ji2019building} provides bi-temporal aerial imagery of Christchurch, New Zealand, with building-change annotations. SECOND \cite{yang2022asymmetric} consists of 4,662 image pairs of $512\times512$ pixels collected from multiple platforms and sensors, covering six land-cover categories. SC-SCD \cite{tan2025triples} contains 1,722 image pairs of $512\times512$ pixels at 0.5 m resolution. Its images were acquired in 2014 and 2016 and provide complete pixel-level semantic annotations for seven land-cover categories.

\subsection{Experimental Setup}

\noindent\textbf{Metrics.}
We evaluate change masks for the queried categories using pixel-level precision, recall, intersection over union (IoU), and F1 score. For SECOND and SC-SCD, we report per-category IoU and F1, together with their averages across categories (mIoU and mF1). All scores are reported as percentages.

\noindent\textbf{Implementation details.}
Our method is implemented in PyTorch, and all experiments are conducted on an NVIDIA GeForce RTX 4090 D GPU. We use pretrained SAM~3 and Qwen2.5-VL-7B-Instruct, with all model parameters frozen and no additional training. Text queries are constructed from dataset category names, while a fixed instruction template guides the MLLM to review ambiguous predictions. Acquired evidence is reused within each image pair. Final change masks are obtained by binarizing the change scores with threshold $\tau$.

\noindent\textbf{Compared methods.}
We compare LookBackOVCD with seven OVCD methods: I-M-C and M-C-I \cite{li2026dynamicearth}, OmniOVCD \cite{zhang2026omniovcd}, AdaptOVCD \cite{dou2026adaptovcd}, OpenDPR \cite{Guo_2026_CVPR}, CoRegOVCD \cite{tang2026coregovcd}, and Seg2Change \cite{su2026seg2change}. For DynamicEarth, I-M-C uses APE for category identification and mask generation, with DINOv2 for temporal comparison. M-C-I uses SAM~2 for mask generation, DINOv2 for temporal comparison, and SegEarth-OV for category identification.

\begin{table}[t]
\centering
\caption{Comparison of OVCD methods on the building change detection datasets LEVIR-CD and WHU-CD.}
\label{tab:levir_whu}
\setlength{\tabcolsep}{5pt}
\renewcommand{\arraystretch}{1.15}
\resizebox{\columnwidth}{!}{
\begin{tabular}{l|cc|cc}
\toprule
\multirow{2}{*}{\textbf{Method}}
& \multicolumn{2}{c|}{\textbf{LEVIR-CD}}
& \multicolumn{2}{c}{\textbf{WHU-CD}} \\
& $\mathrm{IoU}_{C}$ & $\mathrm{F1}_{C}$
& $\mathrm{IoU}_{C}$ & $\mathrm{F1}_{C}$ \\
\midrule

I-M-C \cite{li2026dynamicearth}
& 50.0 & 66.7 & 61.1 & 75.8 \\

M-C-I \cite{li2026dynamicearth}
& 33.8 & 50.5 & 40.9 & 58.1 \\

OmniOVCD \cite{zhang2026omniovcd}
& 67.2 & 80.4 & 66.5 & 79.9 \\

AdaptOVCD \cite{dou2026adaptovcd}
& 51.5 & 68.0 & 62.0 & 76.5 \\

OpenDPR \cite{Guo_2026_CVPR}
& 44.8 & 61.9 & 54.3 & 70.4 \\

CoRegOVCD \cite{tang2026coregovcd}
& \underline{71.0} & \underline{83.0}
& \underline{69.7} & \underline{82.1} \\

% Verify the source of the WHU-CD results before submission.
Seg2Change \cite{su2026seg2change}
& 64.9 & 78.7 & 68.1 & 81.0 \\

\midrule
\rowcolor{oursgray}
\textbf{LookBackOVCD}
& \textbf{75.2} & \textbf{85.8}
& \textbf{73.4} & \textbf{84.7} \\

\bottomrule
\end{tabular}
}
\end{table}

\begin{table*}[t]
\centering
\caption{Comparison of OVCD methods on the semantic change detection dataset SECOND.}
\label{tab:second}
\setlength{\tabcolsep}{4pt}
\renewcommand{\arraystretch}{1.15}
\resizebox{\textwidth}{!}{
\begin{tabular}{l|cc|cc|cc|cc|cc|cc|cc}
\toprule
\multirow{2}{*}{\textbf{Method}}
& \multicolumn{2}{c|}{\textbf{Building}}
& \multicolumn{2}{c|}{\textbf{Tree}}
& \multicolumn{2}{c|}{\textbf{Water}}
& \multicolumn{2}{c|}{\textbf{Low vegetation}}
& \multicolumn{2}{c|}{\textbf{Surface}}
& \multicolumn{2}{c|}{\textbf{Playground}}
& \multicolumn{2}{c}{\textbf{Class Avg.}} \\
& $\mathrm{IoU}_{C}$ & $\mathrm{F1}_{C}$
& $\mathrm{IoU}_{C}$ & $\mathrm{F1}_{C}$
& $\mathrm{IoU}_{C}$ & $\mathrm{F1}_{C}$
& $\mathrm{IoU}_{C}$ & $\mathrm{F1}_{C}$
& $\mathrm{IoU}_{C}$ & $\mathrm{F1}_{C}$
& $\mathrm{IoU}_{C}$ & $\mathrm{F1}_{C}$
& $\mathrm{mIoU}_{C}$ & $\mathrm{mF1}_{C}$ \\
\midrule

I-M-C \cite{li2026dynamicearth}
& 28.1 & 43.9
& 14.1 & 24.8
& 12.2 & 21.7
& 1.4 & 2.7
& 0.0 & 0.0
& 16.0 & 27.6
& 12.0 & 20.1 \\

M-C-I \cite{li2026dynamicearth}
& 36.6 & 53.5
& 18.2 & 30.8
& 13.8 & 24.3
& 22.1 & 36.2
& 19.2 & 32.3
& 17.1 & 29.2
& 21.2 & 34.4 \\

OmniOVCD \cite{zhang2026omniovcd}
& 45.2 & 62.3
& 16.7 & 28.6
& 21.2 & 35.0
& 24.5 & 39.3
& 27.7 & 43.4
& 27.0 & 42.4
& 27.1 & 41.8 \\

AdaptOVCD \cite{dou2026adaptovcd}
& 46.9 & 63.8
& 12.7 & 22.5
& 23.3 & 37.8
& 22.2 & 36.3
& \underline{34.0} & \underline{50.7}
& 29.3 & 45.3
& 28.1 & 42.7 \\

OpenDPR \cite{Guo_2026_CVPR}
& 42.4 & 59.5
& \underline{20.9} & 34.5
& 17.5 & 29.8
& 23.0 & 37.4
& 30.2 & 46.4
& \underline{37.3} & \underline{54.3}
& 28.6 & 43.7 \\

CoRegOVCD \cite{tang2026coregovcd}
& 48.9 & 65.7
& \underline{20.9} & \underline{34.6}
& \textbf{29.9} & \textbf{46.0}
& \underline{27.7} & \underline{43.4}
& 32.3 & 48.8
& 30.3 & 46.5
& \underline{31.7} & \underline{47.5} \\

Seg2Change \cite{su2026seg2change}
& \textbf{61.8} & \textbf{76.4}
& 16.1 & 27.7
& 22.9 & 37.2
& 18.7 & 31.5
& \textbf{37.3} & \textbf{54.3}
& 17.8 & 30.2
& 29.1 & 42.9 \\

\midrule
\rowcolor{oursgray}
\textbf{LookBackOVCD}
& \underline{55.1} & \underline{71.1}
& \textbf{22.6} & \textbf{36.9}
& \underline{27.8} & \underline{43.5}
& \textbf{30.1} & \textbf{46.3}
& 32.9 & 49.5
& \textbf{40.6} & \textbf{57.8}
& \textbf{34.9} & \textbf{50.9} \\

\bottomrule
\end{tabular}
}
\end{table*}

\providecolor{oursgray}{gray}{0.93}

\begin{table*}[t]
\centering
\caption{Comparison of OVCD methods on the semantic change detection dataset SC-SCD.}
\label{tab:sc_scd}
\setlength{\tabcolsep}{4pt}
\renewcommand{\arraystretch}{1.15}
\resizebox{\textwidth}{!}{%
\begin{tabular}{l|cc|cc|cc|cc|cc|cc|cc|cc}
\toprule
\multirow{2}{*}{\textbf{Method}}
& \multicolumn{2}{c|}{\textbf{Bareland}}
& \multicolumn{2}{c|}{\textbf{Water}}
& \multicolumn{2}{c|}{\textbf{Building}}
& \multicolumn{2}{c|}{\textbf{Structure}}
& \multicolumn{2}{c|}{\textbf{Farmland}}
& \multicolumn{2}{c|}{\textbf{Vegetation}}
& \multicolumn{2}{c|}{\textbf{Road}}
& \multicolumn{2}{c}{\textbf{Class Avg.}} \\
& $\mathrm{IoU}_{C}$ & $\mathrm{F1}_{C}$
& $\mathrm{IoU}_{C}$ & $\mathrm{F1}_{C}$
& $\mathrm{IoU}_{C}$ & $\mathrm{F1}_{C}$
& $\mathrm{IoU}_{C}$ & $\mathrm{F1}_{C}$
& $\mathrm{IoU}_{C}$ & $\mathrm{F1}_{C}$
& $\mathrm{IoU}_{C}$ & $\mathrm{F1}_{C}$
& $\mathrm{IoU}_{C}$ & $\mathrm{F1}_{C}$
& $\mathrm{mIoU}_{C}$ & $\mathrm{mF1}_{C}$ \\
\midrule

% DynamicEarth results reported in Table 15 of Seg2Change.
I-M-C \cite{li2026dynamicearth}
& 0.0 & 0.0
& 12.3 & 22.0
& 13.6 & 23.9
& 0.0 & 0.0
& 0.0 & 0.0
& 0.0 & 0.0
& 10.2 & 18.5
& 5.2 & 9.2 \\

M-C-I \cite{li2026dynamicearth}
& 22.6 & 36.9
& 33.1 & 49.8
& \underline{28.6} & \underline{44.4}
& \underline{13.2} & \underline{23.3}
& 5.5 & 10.4
& 19.8 & 33.0
& 3.1 & 6.0
& 18.0 & 29.1 \\

OmniOVCD \cite{zhang2026omniovcd}
& 26.7 & 42.1
& 42.3 & 59.5
& 19.8 & 33.1
& 9.2 & 16.8
& 11.5 & 20.6
& 30.9 & 47.2
& 12.4 & 22.1
& 21.8 & 34.5 \\

AdaptOVCD \cite{dou2026adaptovcd}
& \underline{29.7} & \underline{45.8}
& 44.8 & 61.9
& 20.1 & 33.5
& 11.5 & 20.6
& 12.0 & 21.4
& 28.1 & 43.9
& 12.3 & 21.9
& 22.6 & 35.6 \\

OpenDPR \cite{Guo_2026_CVPR}
& 28.3 & 44.1
& 38.2 & 55.3
& 17.9 & 30.4
& 10.2 & 18.5
& 11.8 & 21.1
& 30.4 & 46.6
& 15.7 & 27.1
& 21.8 & 34.7 \\

CoRegOVCD \cite{tang2026coregovcd}
& 28.8 & 44.7
& \underline{46.3} & \underline{63.3}
& 22.1 & 36.2
& 12.5 & 22.2
& \underline{12.2} & \underline{21.7}
& \underline{33.6} & \underline{50.3}
& 13.3 & 23.5
& \underline{24.1} & \underline{37.4} \\

Seg2Change \cite{su2026seg2change}
& 27.5 & 43.2
& 21.9 & 35.9
& \textbf{47.2} & \textbf{64.1}
& 0.8 & 1.5
& 11.4 & 20.5
& 29.8 & 46.0
& \textbf{23.9} & \textbf{38.6}
& 23.2 & 35.7 \\

\midrule
\rowcolor{oursgray}
\textbf{LookBackOVCD}
& \textbf{31.9} & \textbf{48.4}
& \textbf{49.8} & \textbf{66.5}
& 26.2 & 41.5
& \textbf{15.6} & \textbf{27.0}
& \textbf{15.5} & \textbf{26.8}
& \textbf{36.7} & \textbf{53.7}
& \underline{15.9} & \underline{27.4}
& \textbf{27.4} & \textbf{41.6} \\

\bottomrule
\end{tabular}%
}
\end{table*}
\subsection{Comparison with State-of-the-Art Methods}

\noindent\textbf{Results on Building Change Detection.} Table~\ref{tab:levir_whu} compares LookBackOVCD with existing OVCD methods on LEVIR-CD and WHU-CD. Without additional training, our method achieves the highest $\mathrm{IoU}_{C}$ and $\mathrm{F1}_{C}$ among the compared methods on both datasets. On LEVIR-CD, LookBackOVCD reaches 75.2\% IoU and 85.8\% F1, exceeding CoRegOVCD by 4.2 and 2.8 percentage points, respectively. On WHU-CD, it obtains 73.4\% IoU and 84.7\% F1, improving over the same baseline by 3.7 and 2.6 points. These gains suggest that MLLM-based review of target category and mask coverage in both images can provide useful feedback for refining predictions before temporal comparison.

\noindent\textbf{Results on Semantic Change Detection.}
Tables~\ref{tab:second} and~\ref{tab:sc_scd} report the semantic change detection results on SECOND and SC-SCD. LookBackOVCD achieves the highest class-average IoU and F1 among the compared methods on both datasets. On SECOND, it obtains 34.9\% $\mathrm{mIoU}_{C}$ and 50.9\% $\mathrm{mF1}_{C}$, exceeding CoRegOVCD by 3.2 and 3.4 percentage points, respectively. On SC-SCD, it reaches 27.4\% $\mathrm{mIoU}_{C}$ and 41.6\% $\mathrm{mF1}_{C}$, with corresponding gains of 3.3 and 4.2 points. At the category level, our method performs best on tree, low vegetation, and playground in SECOND, and on five of the seven categories in SC-SCD, including water, structure, and vegetation. For example, on playground in SECOND, LookBackOVCD improves IoU from 37.3\% to 40.6\% and F1 from 54.3\% to 57.8\% over OpenDPR. However, Seg2Change performs better on building in both datasets and on road in SC-SCD, while CoRegOVCD remains stronger on water in SECOND. These results suggest that MLLM feedback can improve target predictions across diverse land-cover categories, although the benefits vary across classes.

\subsection{Ablation Studies and Analyses}
\noindent\textbf{Component ablation study.}
Table~\ref{tab:component_ablation} evaluates MLLM feedback and the shared evidence pool. The baseline compares initial SAM3 semantic responses without feedback. Without the evidence pool, MLLM feedback raises IoU from 53.1\% to 73.7\% on LEVIR-CD and from 50.5\% to 72.1\% on WHU-CD, with consistent gains on SECOND and SC-SCD. These results support revisiting target category and mask coverage through visual review and feedback before temporal comparison. The evidence pool further improves both metrics across all four datasets, suggesting that evidence reuse benefits subsequent judgments.

\begin{table*}[t]
\centering
\caption{Ablation study of different components in LookBackOVCD.}
\label{tab:component_ablation}

\small
\setlength{\tabcolsep}{5pt}
\setlength{\arrayrulewidth}{0.5pt}
\renewcommand{\arraystretch}{1.2}

\begin{tabularx}{0.94\textwidth}{
    l|
    *{2}{>{\centering\arraybackslash}X}|
    *{2}{>{\centering\arraybackslash}X}|
    *{2}{>{\centering\arraybackslash}X}|
    *{2}{>{\centering\arraybackslash}X}
}
\specialrule{0.8pt}{0pt}{0pt}
\multirow{2}{*}{\textbf{Setting}}
& \multicolumn{2}{c|}{\textbf{LEVIR-CD}}
& \multicolumn{2}{c|}{\textbf{WHU-CD}}
& \multicolumn{2}{c|}{\textbf{SECOND}}
& \multicolumn{2}{c}{\textbf{SC-SCD}} \\
\cline{2-9}
& $\mathrm{IoU}_{C}$ & $\mathrm{F1}_{C}$
& $\mathrm{IoU}_{C}$ & $\mathrm{F1}_{C}$
& $\mathrm{mIoU}_{C}$ & $\mathrm{mF1}_{C}$
& $\mathrm{mIoU}_{C}$ & $\mathrm{mF1}_{C}$ \\
\hline

Baseline
& 53.1 & 69.4
& 50.5 & 67.1
& 24.3 & 38.4
& 18.3 & 29.4 \\

w/o evidence pool
& 73.7 & 84.9
& 72.1 & 83.8
& 34.0 & 49.8
& 26.3 & 40.3 \\

\rowcolor{oursgray}
\textbf{LookBackOVCD}
& 75.2 & 85.8
& 73.4 & 84.7
& 34.9 & 50.9
& 27.4 & 41.6 \\

\ablationwhiterow
\specialrule{0.8pt}{0pt}{0pt}
\end{tabularx}
\end{table*}

\begin{table}[t]
\centering
\caption{Ablation study of local review prompts.}
\label{tab:prompt_ablation}

\small
\setlength{\tabcolsep}{4pt}
\setlength{\arrayrulewidth}{0.5pt}
\renewcommand{\arraystretch}{1.2}

\begin{tabularx}{\columnwidth}{
    l|
    *{2}{>{\centering\arraybackslash}X}|
    *{2}{>{\centering\arraybackslash}X}
}
\specialrule{0.8pt}{0pt}{0pt}
\multirow{2}{*}{\textbf{Prompt}}
& \multicolumn{2}{c|}{\textbf{LEVIR-CD}}
& \multicolumn{2}{c}{\textbf{SECOND}} \\
\cline{2-5}
& $\mathrm{IoU}_{C}$ & $\mathrm{F1}_{C}$
& $\mathrm{mIoU}_{C}$ & $\mathrm{mF1}_{C}$ \\
\hline

Prompt 1
& 70.1 & 82.4
& 30.1 & 45.4 \\

Prompt 2
& 71.6 & 83.4
& 31.4 & 46.8 \\

\multicolumn{1}{c|}{\textbf{Ours}}
& 75.2 & 85.8
& 34.9 & 50.9 \\

\specialrule{0.8pt}{0pt}{0pt}
\end{tabularx}
\end{table}
\noindent\textbf{Local review prompts.}
Table~\ref{tab:prompt_ablation} compares three local review prompts
on LEVIR-CD and SECOND. Prompt 1 specifies only the review task,
while Prompt 2 additionally requests a brief visual description
without specifying what to examine. Our prompt guides the model
to assess target category and mask coverage using object attributes
and surrounding context. Prompt 2 improves both metrics over Prompt 1
on both datasets. Compared with Prompt 2, our prompt increases IoU
by 3.6 percentage points on LEVIR-CD and mIoU by 3.5 points on SECOND.
These results indicate that directing visual observations toward
category and coverage errors provides more useful feedback for
correcting predictions before temporal comparison.

\noindent\textbf{MLLM backbones.}
Table~\ref{tab:mllm_backbones} compares four MLLM backbones within LookBackOVCD. Qwen2.5-VL-7B achieves the highest mIoU and mF1 of 34.9\% and 50.9\%, respectively, with the lowest measured latency of 13.5 s/pair. InternVL3-2B trails by only 0.6 percentage points in both metrics while reducing peak memory from 19.9 to 8.8 GiB and maintaining similar inference speed. Qwen2.5-VL-3B also reduces memory consumption, but shows a larger accuracy drop without a latency benefit. LLaVA-OneVision-7B requires 42.5 s/pair while achieving lower accuracy than Qwen2.5-VL-7B. These results support our choice of Qwen2.5-VL-7B as the default backbone, with InternVL3-2B providing a competitive alternative when GPU memory is limited.

\begin{table}[t]
\centering
\caption{Ablation of MLLM backbones on SECOND. Time and peak memory are measured on SECOND (building) using an NVIDIA GeForce RTX 4090 D GPU.}
\label{tab:mllm_backbones}

\small
\setlength{\tabcolsep}{2pt}
\renewcommand{\arraystretch}{1.15}

\begin{tabular*}{\columnwidth}{
    @{\extracolsep{\fill}}lcccc@{}
}
\toprule
MLLM backbone
& mIoU
& mF1
& \makecell{Time\\(s/pair)}
& \makecell{Memory\\(GiB)} \\
\midrule
InternVL3-2B~\cite{zhu2025internvl3}
& 34.3 & 50.3 & 13.7 & 8.8 \\
Qwen2.5-VL-3B~\cite{bai2025qwen25vl}
& 33.4 & 49.4 & 14.9 & 11.2 \\
LLaVA-OneVision-7B~\cite{li2025llavaonevision}
& 34.1 & 50.2 & 42.5 & 20.2 \\
Qwen2.5-VL-7B~\cite{bai2025qwen25vl}
& 34.9 & 50.9 & 13.5 & 19.9 \\
\bottomrule
\end{tabular*}

\end{table}

\begin{figure}[t]
    \centering
    \includegraphics[width=\columnwidth]{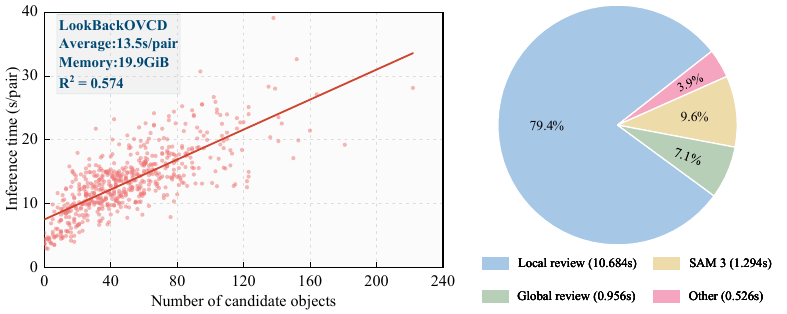}
    \caption{Inference time with different numbers of candidate objects and time distribution across stages on SECOND (building), measured on an NVIDIA GeForce RTX 4090 D GPU.}
    \label{fig:efficiency}
\end{figure}

\noindent\textbf{Efficiency analysis.}
Figure~\ref{fig:efficiency} shows inference time and its distribution across stages on SECOND (building). LookBackOVCD takes 13.5 s per image pair on average, with a peak GPU memory usage of 19.9 GiB. Inference time generally increases with the number of candidate objects. Local review accounts for 79.4\% of total time, including visual observation and feedback generation. Global review and SAM~3 account for 7.1\% and 9.6\%, respectively, while the remaining operations take 3.9\%. Local MLLM review is therefore the main runtime bottleneck, particularly in scenes containing many candidate objects.

\subsection{Visualization Results}
\begin{figure*}[t]
    \centering
    \includegraphics[width=\textwidth]{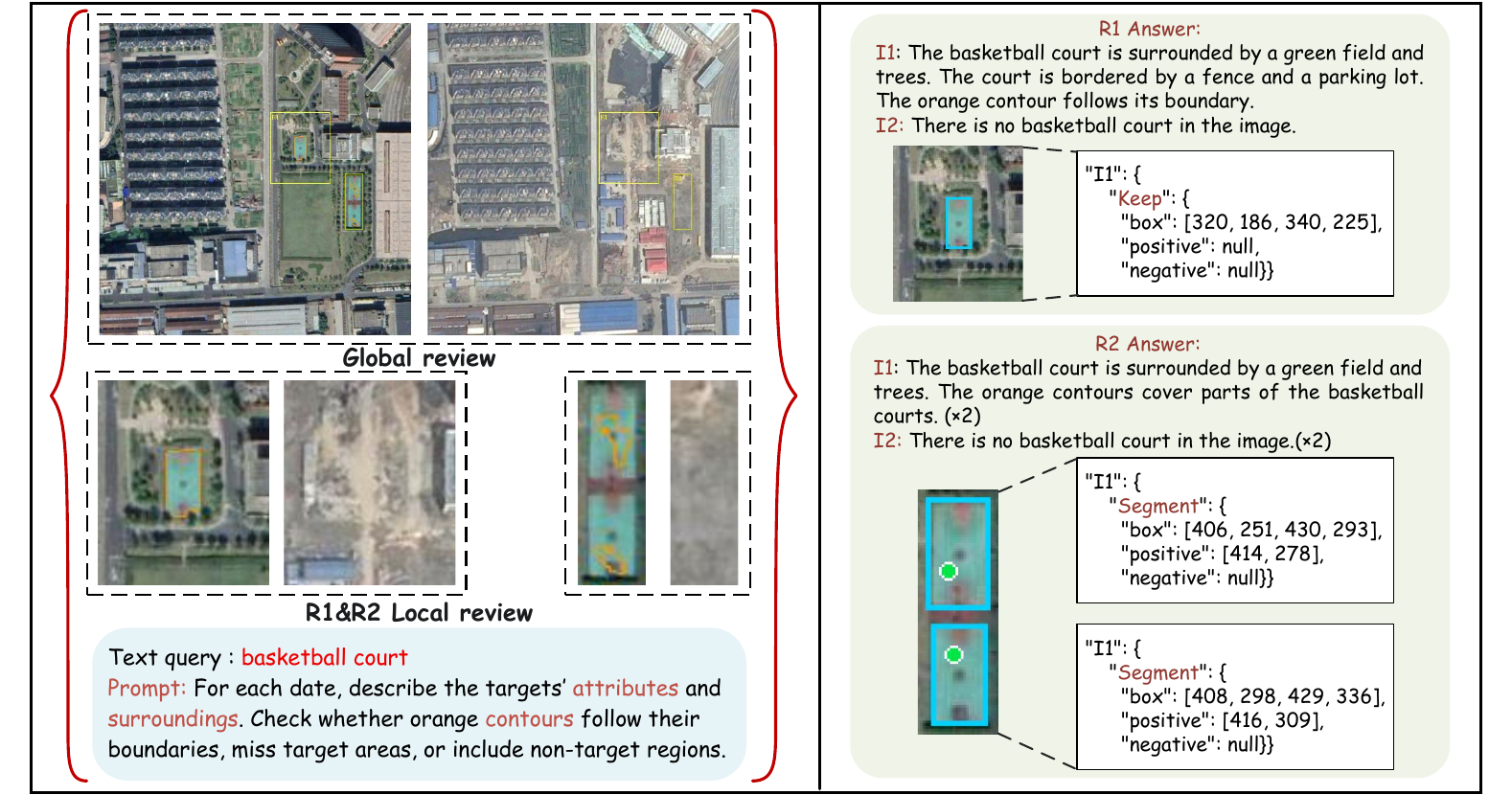}
    \caption{Illustration of the MLLM feedback process.}
    \label{fig:feedback_process}
\end{figure*}

\noindent\textbf{Feedback process.}
Figure~\ref{fig:feedback_process} illustrates how MLLM feedback guides corrections before temporal comparison. Target attributes and surroundings support category judgments, while comparing the orange contours with visible boundaries reveals missing or excess coverage. These observations guide the spatial prompts passed to SAM~3. In R1, agreement between the prediction and visible target leads to Keep, which reuses the existing object box without point prompts. In R2, Segment uses object boxes and positive points to correct incomplete coverage of the two adjacent courts within the same review region. Boxes delimit individual targets, while points indicate areas requiring correction. SAM~3 uses these prompts to refine predictions in each image for subsequent temporal comparison.

\begin{figure}[t]
    \centering
    \includegraphics[width=\columnwidth]{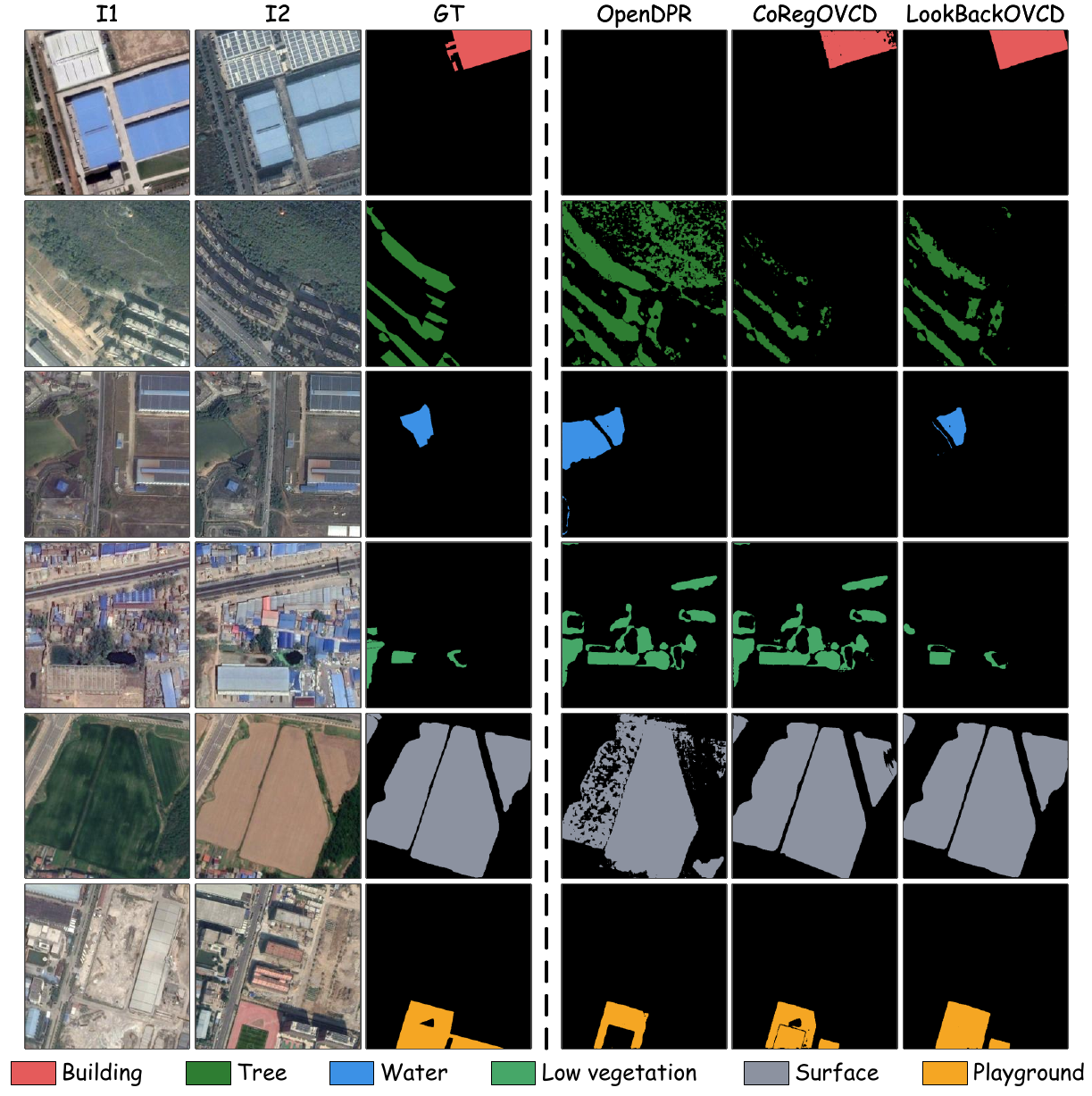}
    \caption{Qualitative comparison on SECOND.}
    \label{fig:qualitative_second}
\end{figure}

\noindent\textbf{Qualitative comparison.}
Figure~\ref{fig:qualitative_second} compares LookBackOVCD with OpenDPR and CoRegOVCD on SECOND. Our method produces change masks that generally agree more closely with the ground truth, with fewer missed regions and spurious responses. These observations support using MLLM feedback to refine target predictions by reassessing category membership and mask coverage in context before temporal comparison.
\section{Conclusion}
\label{sec:conclusion}

We propose LookBackOVCD, a training-free OVCD framework that uses MLLM feedback to correct target predictions before temporal comparison. The framework combines global review of both original images with local review of regions requiring further verification. By examining object attributes, surrounding context, and predicted masks, the MLLM reassesses target category and mask coverage and identifies potentially overlooked targets. Its feedback guides local corrections to semantic responses and prompts SAM~3 to generate or refine masks in each image. A shared evidence pool retains observations within each image pair for reference in subsequent judgments. Experiments on LEVIR-CD, WHU-CD, SECOND, and SC-SCD demonstrate that LookBackOVCD outperforms existing training-free methods in both building and multi-class change detection. These results support using visual verification and feedback to reduce the impact of initial recognition errors on change detection. The separation of visual review and mask correction allows the framework to incorporate advances in both MLLMs and segmentation models. Future work will explore more efficient local review and improved localization of small targets in dense scenes.

\section*{LLM Usage Statement}
We used OpenAI ChatGPT and Codex to assist with research and code
development, experimental analysis, and the preparation and revision
of manuscript text and figures. The authors take full responsibility
for all submitted content.

{
    \small
    \bibliographystyle{ieeenat_fullname}
    \bibliography{main}
}

\clearpage
\maketitlesupplementary
\appendix
\section*{Supplementary Material Contents}

\begin{itemize}[
    leftmargin=*,
    itemsep=0pt,
    parsep=0pt,
    topsep=4pt
]
    \item \textbf{Section~\ref{sec:supp_experimental_details}
    -- More Experimental Details.}

    \item \textbf{Section~\ref{sec:supp_method}
    -- Additional Method Details.}
    \item \textbf{Section~\ref{sec:supp_experiments}
-- More Diagnostic Experiments.}
    \item \textbf{Section~\ref{sec:supp_qualitative} -- More Qualitative Results.}
\end{itemize}

\section{More Experimental Details}
\label{sec:supp_experimental_details}

\subsection{Dataset Description}
\label{sec:supp_datasets}

\noindent\textbf{LEVIR-CD.}
LEVIR-CD~\cite{chen2020spatial} is a building change detection
dataset containing 637 pairs of Google Earth images collected
from 20 regions in Texas, USA. Each image has a size of
$1024\times1024$ pixels and a spatial resolution of 0.5 m.
The image pairs span 5--14 years and cover various building
types, including residential buildings, garages, and warehouses.
Binary annotations indicate building construction and demolition.
We crop the 128 test image pairs into non-overlapping
$256\times256$ patches and evaluate our method on all 2,048
resulting patch pairs.

\noindent\textbf{WHU-CD.}
WHU-CD~\cite{ji2019building} provides bi-temporal aerial
imagery of Christchurch, New Zealand, acquired in 2012 and
2016, together with pixel-level building change annotations.
It covers building changes across residential, commercial,
and industrial areas. We use the version containing 7,434
image pairs of $256\times256$ pixels, with a training,
validation, and test split of approximately $8:1:1$.
Following the test partition adopted by
DynamicEarth~\cite{li2026dynamicearth}, we perform inference
on all 744 test pairs.

\noindent\textbf{SECOND.}
SECOND~\cite{yang2022asymmetric} is a semantic change detection
dataset containing 4,662 image pairs collected from multiple
sensors over cities in China, including Hangzhou, Chengdu,
and Shanghai. The images have a size of $512\times512$ pixels
and spatial resolutions ranging from 0.5 to 3 m.
The annotations cover six land-cover categories: building,
tree, water, low vegetation, non-vegetated ground surface,
and playground. We follow the evaluation partition used by
DynamicEarth~\cite{li2026dynamicearth} and evaluate all 593
image pairs specified in the file list released by its authors.
Each pair is processed at its original image size for all
six categories.

\noindent\textbf{SC-SCD.}
SC-SCD~\cite{tan2025triples} provides bi-temporal satellite
imagery acquired in 2014 and 2016 at a spatial resolution
of 0.5 m. It contains 1,722 image pairs of $512\times512$
pixels, divided into 1,400 training pairs and 322 test pairs.
The dataset provides semantic annotations for both images
and corresponding binary change maps. Its seven land-cover
categories are bareland, water, building, structure, farmland,
vegetation, and road. We evaluate all 322 test pairs at their
original image size for each of the seven categories.

\subsection{Evaluation Metrics}
\label{sec:supp_metrics}

We evaluate the predicted change masks using pixel-level
precision, recall, intersection over union (IoU), and F1 score.
For each queried category, true positives (TP) count correctly
detected changed pixels, false positives (FP) count pixels
incorrectly predicted as changes of that category, and false
negatives (FN) count missed changed pixels. These counts are
accumulated over all evaluated image pairs before computing
the metrics.

Precision measures the proportion of predicted changed pixels
that are correct, while recall measures the proportion of
ground-truth changed pixels that are detected:
\begin{equation}
\mathrm{Pre}_{C}
=
\frac{\mathrm{TP}}{\mathrm{TP}+\mathrm{FP}},
\label{eq:supp_precision}
\end{equation}
\begin{equation}
\mathrm{Rec}_{C}
=
\frac{\mathrm{TP}}{\mathrm{TP}+\mathrm{FN}}.
\label{eq:supp_recall}
\end{equation}
Here, the subscript $C$ denotes the change foreground.
IoU measures the overlap between the predicted and
ground-truth change regions:
\begin{equation}
\mathrm{IoU}_{C}
=
\frac{\mathrm{TP}}
{\mathrm{TP}+\mathrm{FP}+\mathrm{FN}}.
\label{eq:supp_iou}
\end{equation}
F1 is the harmonic mean of precision and recall:
\begin{equation}
\mathrm{F1}_{C}
=
\frac{2\mathrm{TP}}
{2\mathrm{TP}+\mathrm{FP}+\mathrm{FN}}.
\label{eq:supp_f1}
\end{equation}

For SECOND and SC-SCD, we compute the metrics separately
for each category and report the arithmetic means of
the category-level IoU and F1 scores:
\begin{equation}
\begin{aligned}
\mathrm{mIoU}_{C}
&=
\frac{1}{K}
\sum_{c=1}^{K}\mathrm{IoU}_{C,c},\\
\mathrm{mF1}_{C}
&=
\frac{1}{K}
\sum_{c=1}^{K}\mathrm{F1}_{C,c},
\end{aligned}
\label{eq:supp_mean_metrics}
\end{equation}
where $\mathrm{IoU}_{C,c}$ and $\mathrm{F1}_{C,c}$ are the
scores for category $c$, and $K$ is the number of evaluated
categories: six for SECOND and seven for SC-SCD.
Each category contributes equally to the reported averages.
All accuracy metrics are expressed as percentages.

We also report average inference time per image pair
in seconds and peak GPU memory usage in GiB.
Both are measured on SECOND (building) using a single
NVIDIA GeForce RTX 4090 D GPU. Inference time includes
initial perception, global and local review, feedback
execution, and the remaining inference operations.
Model loading, warm-up, and metric computation are excluded.
Peak memory is the maximum GPU memory usage observed
during the measured inference.

% 指标定义、公式和统计方式。

\subsection{Implementation Details}
\label{sec:supp_implementation}

\noindent\textbf{Models and environment.}
Our framework is implemented in PyTorch and runs on a single
NVIDIA GeForce RTX 4090 D GPU. We use pretrained SAM~3
for initial perception and mask correction, and
Qwen2.5-VL-7B-Instruct as the default MLLM for visual review.
The MLLM is loaded through the Transformers library.
All foundation models remain frozen throughout inference.

\noindent\textbf{Parameter settings for initial perception.}For each evaluation setting, we manually adjust the instance confidence
threshold and the maximum number of retained instances to obtain a
suitable set of initial candidates for the SAM~3 baseline. Following
the default mask threshold in the official SAM~3 implementation,
we binarize the aggregated response maps at 0.5 to obtain the contours
used for visual review. The baseline and the corresponding variant
with MLLM feedback use the same initial predictions.

\noindent\textbf{Inference pipeline.}
The framework takes a pair of RGB images and a text query
constructed from the target category name.
SAM~3 first produces semantic responses and instance masks
for each image. Qwen2.5-VL jointly examines the full images
and initial predictions, then inspects paired crops from
regions requiring further verification. Each local review
receives the full images, local crops with predicted contours,
object coordinates, and relevant observations retrieved
from the evidence pool.
The model generates a brief visual observation followed
by structured feedback specifying correction actions and
spatial prompts. Boxes and points are mapped from crop
coordinates to the original image coordinates and passed
to SAM~3 for mask correction.
The updated predictions are then compared to produce
the change mask for the queried category.

\subsection{Prompt Templates}
\label{sec:supp_prompts}

\noindent\textbf{SAM~3 prompts.}
Table~\ref{tab:sam3_text_prompts} lists the text prompts
provided to SAM~3 for the evaluated categories.
For each category, the same prompt is applied separately
to both images to obtain the initial target predictions.
The prompts follow the category names or their corresponding
land-cover concepts.

\begin{table}[t]
\centering
\caption{Text prompts provided to SAM~3 for each dataset
and category.}
\label{tab:sam3_text_prompts}

\small
\setlength{\tabcolsep}{4pt}
\renewcommand{\arraystretch}{1.2}

\begin{tabularx}{\columnwidth}{
    >{\centering\arraybackslash}X|
    >{\centering\arraybackslash}X|
    >{\centering\arraybackslash}X
}
\Xhline{0.8pt}
\textbf{Dataset}
& \textbf{Category}
& \textbf{Text prompt} \\
\hline

\textbf{LEVIR-CD}~\cite{chen2020spatial}
& Building & building \\
\hline

\textbf{WHU-CD}~\cite{ji2019building}
& Building & building \\
\hline

\multirow{6}{*}{\textbf{SECOND}~\cite{yang2022asymmetric}}
& Building       & building \\
& Tree           & tree \\
& Water          & water \\
& Low vegetation & grass \\
& Surface        & bare land \\
& Playground     & basketball court \\
\hline

\multirow{7}{*}{\textbf{SC-SCD}~\cite{tan2025triples}}
& Bareland   & bare land \\
& Water      & water \\
& Building   & building \\
& Structure  & structure \\
& Farmland   & farmland \\
& Vegetation & vegetation \\
& Road       & road \\
\Xhline{0.8pt}
\end{tabularx}
\end{table}

\begin{figure*}[tp]
    \centering
    \includegraphics[width=\textwidth]{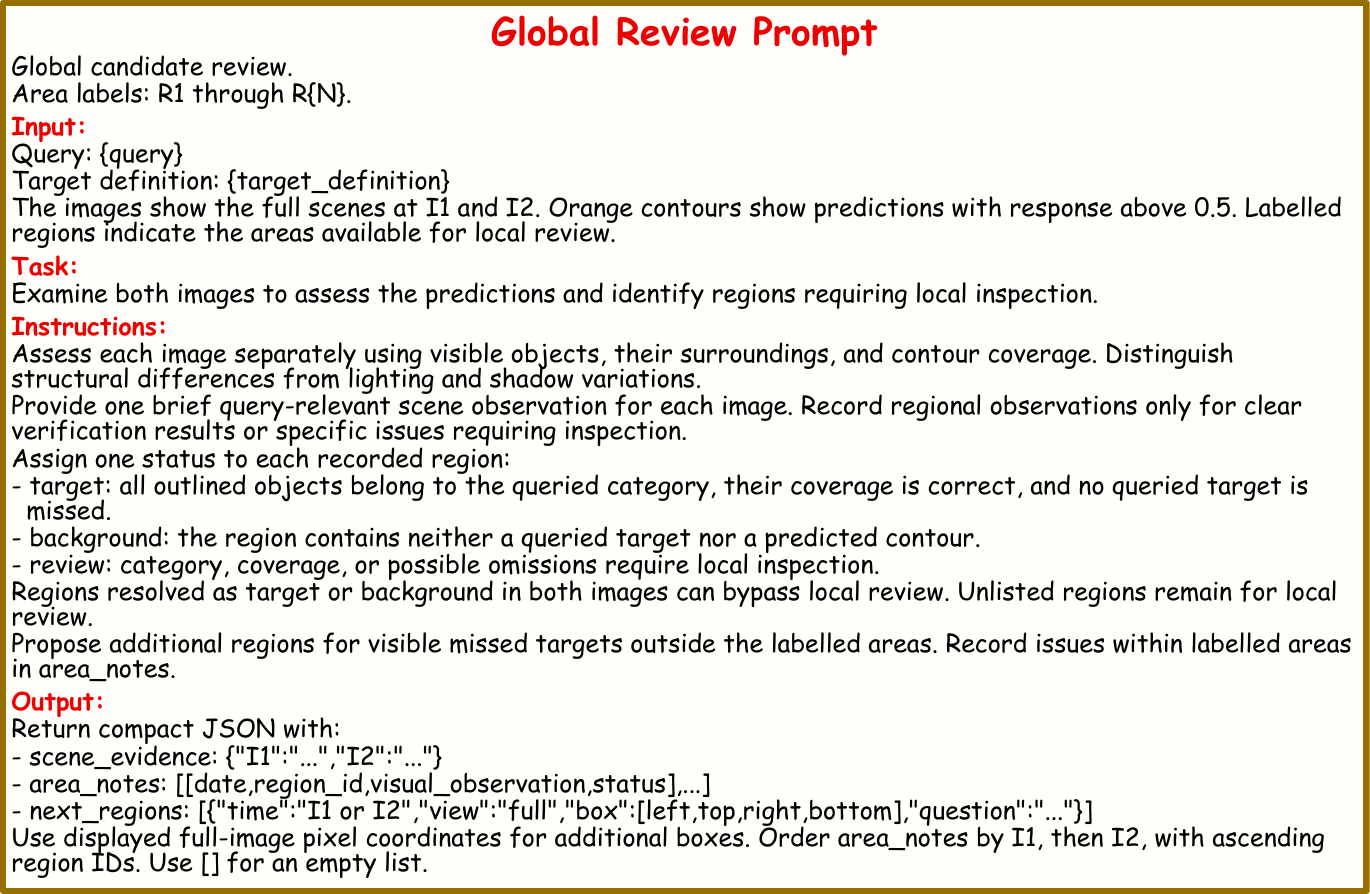}
    \caption{Global review prompt.}
    \label{fig:global_review_prompt}
\end{figure*}

\begin{figure*}[tp]
    \centering
    \includegraphics[width=\textwidth]{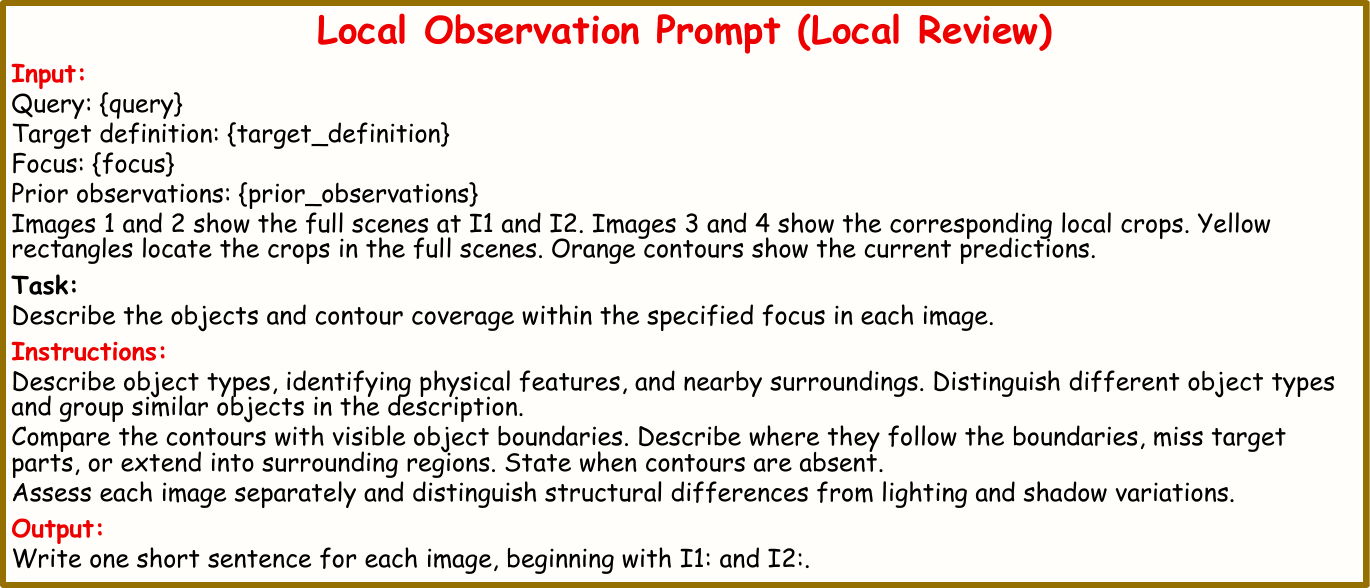}
    \caption{Local observation prompt.}
    \label{fig:local_observation_prompt}
\end{figure*}

\begin{figure*}[tp]
    \centering
    \includegraphics[width=\textwidth]{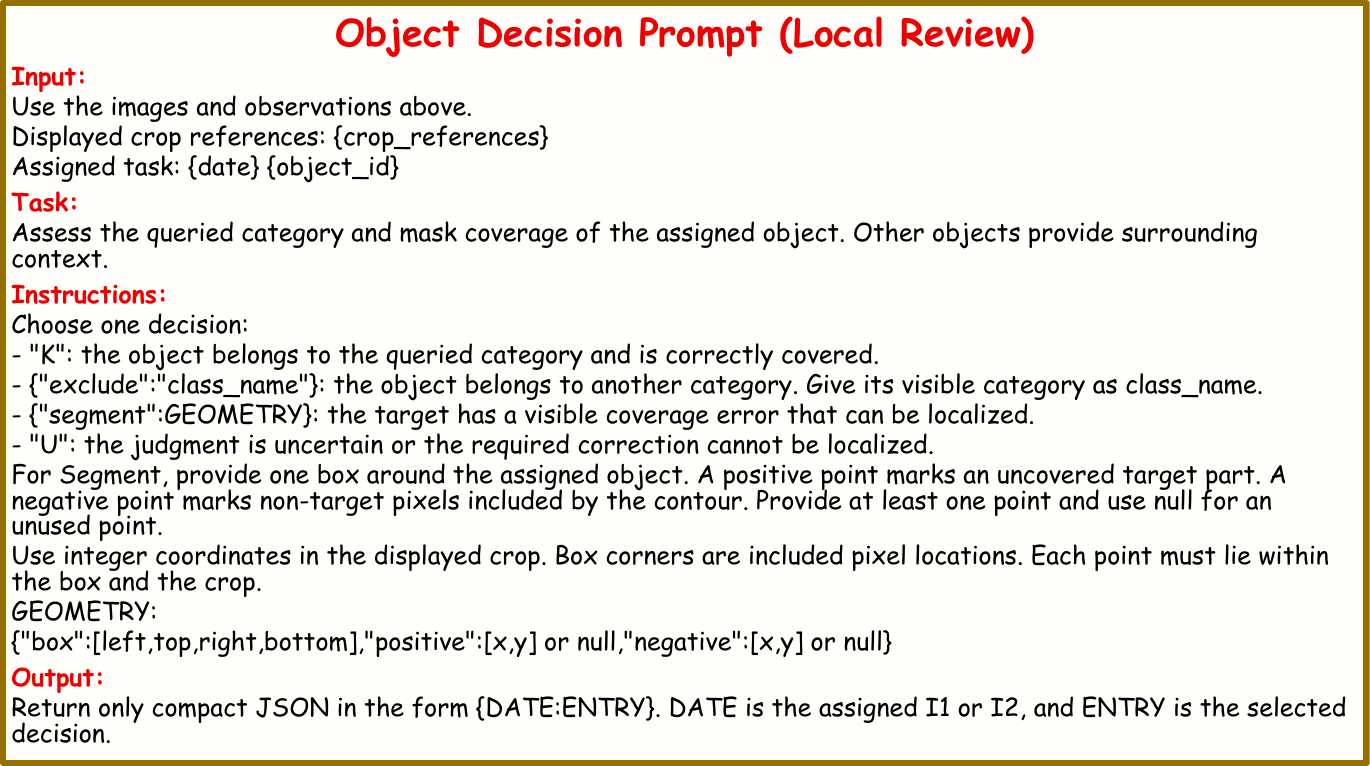}
    \caption{Object decision prompt.}
    \label{fig:object_decision_prompt}
\end{figure*}

\begin{figure*}[tp]
    \centering
    \includegraphics[width=\textwidth]{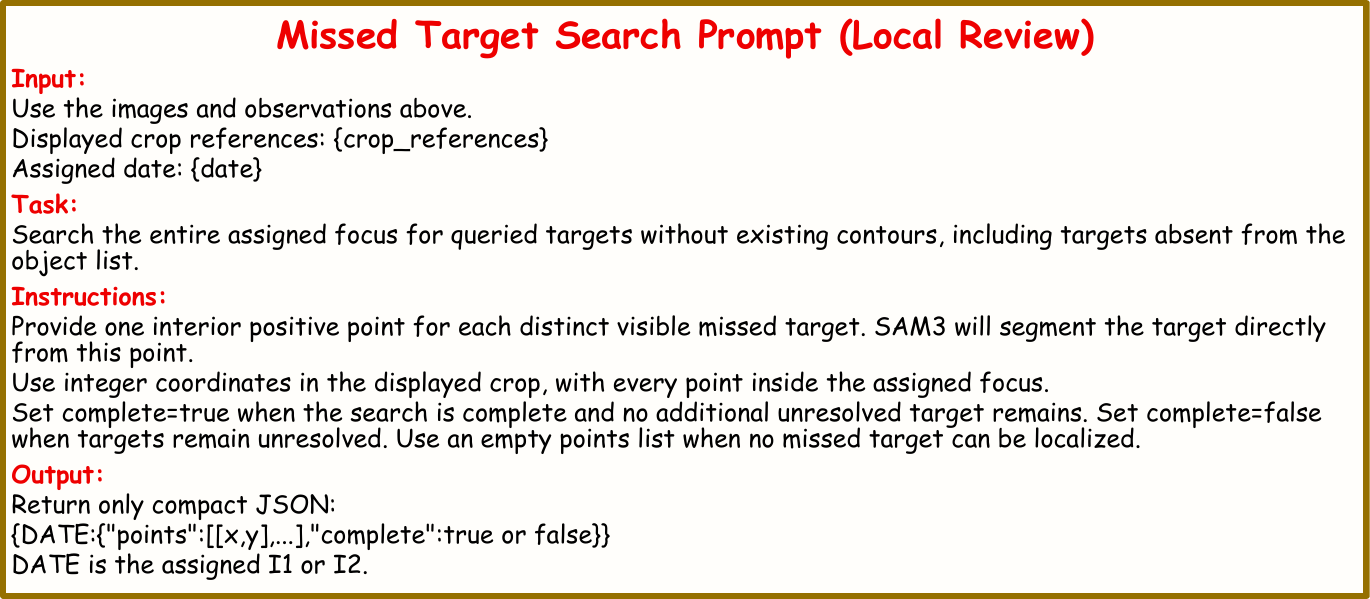}
    \caption{Missed target search prompt.}
    \label{fig:missed_target_search_prompt}
\end{figure*}

\begin{figure*}[tp]
    \centering
    \includegraphics[width=\textwidth]{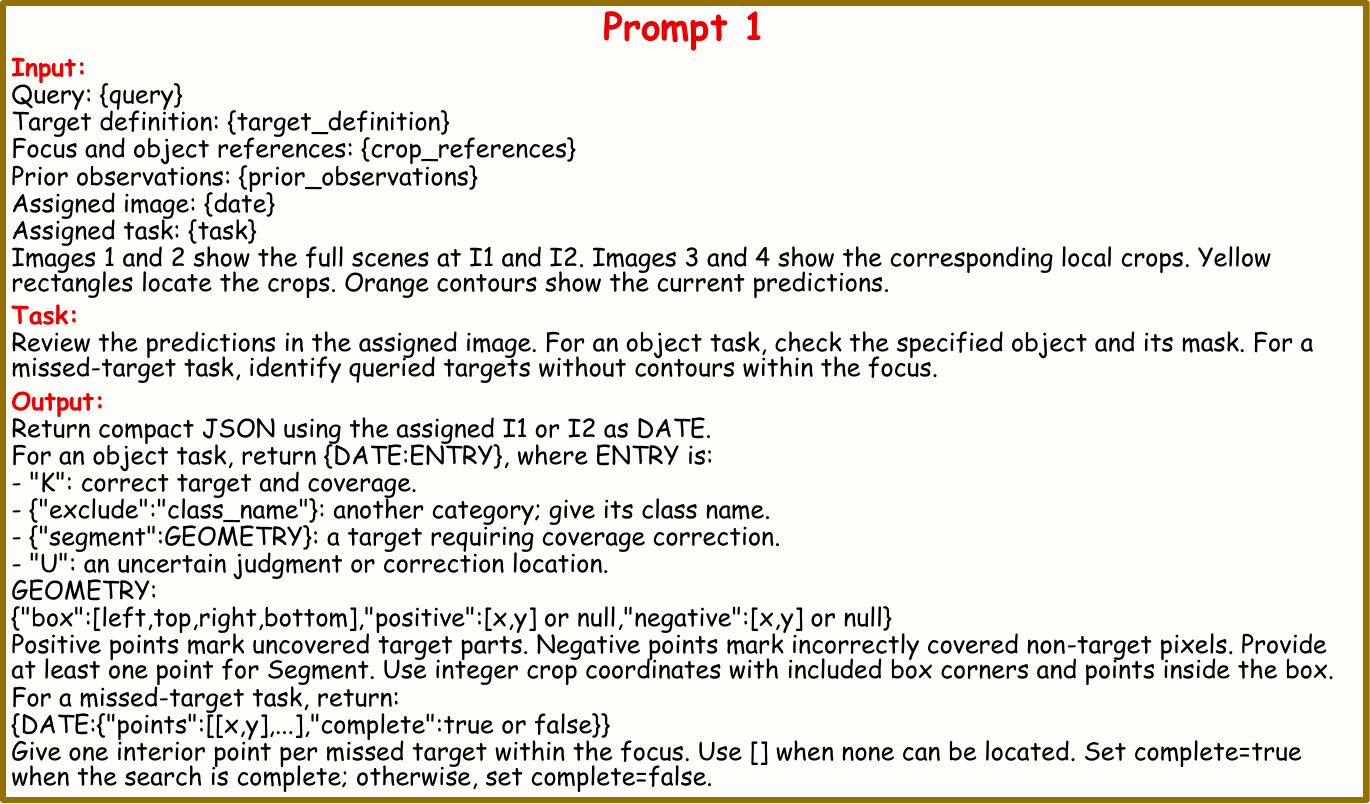}
    \caption{Prompt 1.}
    \label{fig:ablation_prompt1}
\end{figure*}

\begin{figure*}[tp]
    \centering
    \includegraphics[width=\textwidth]{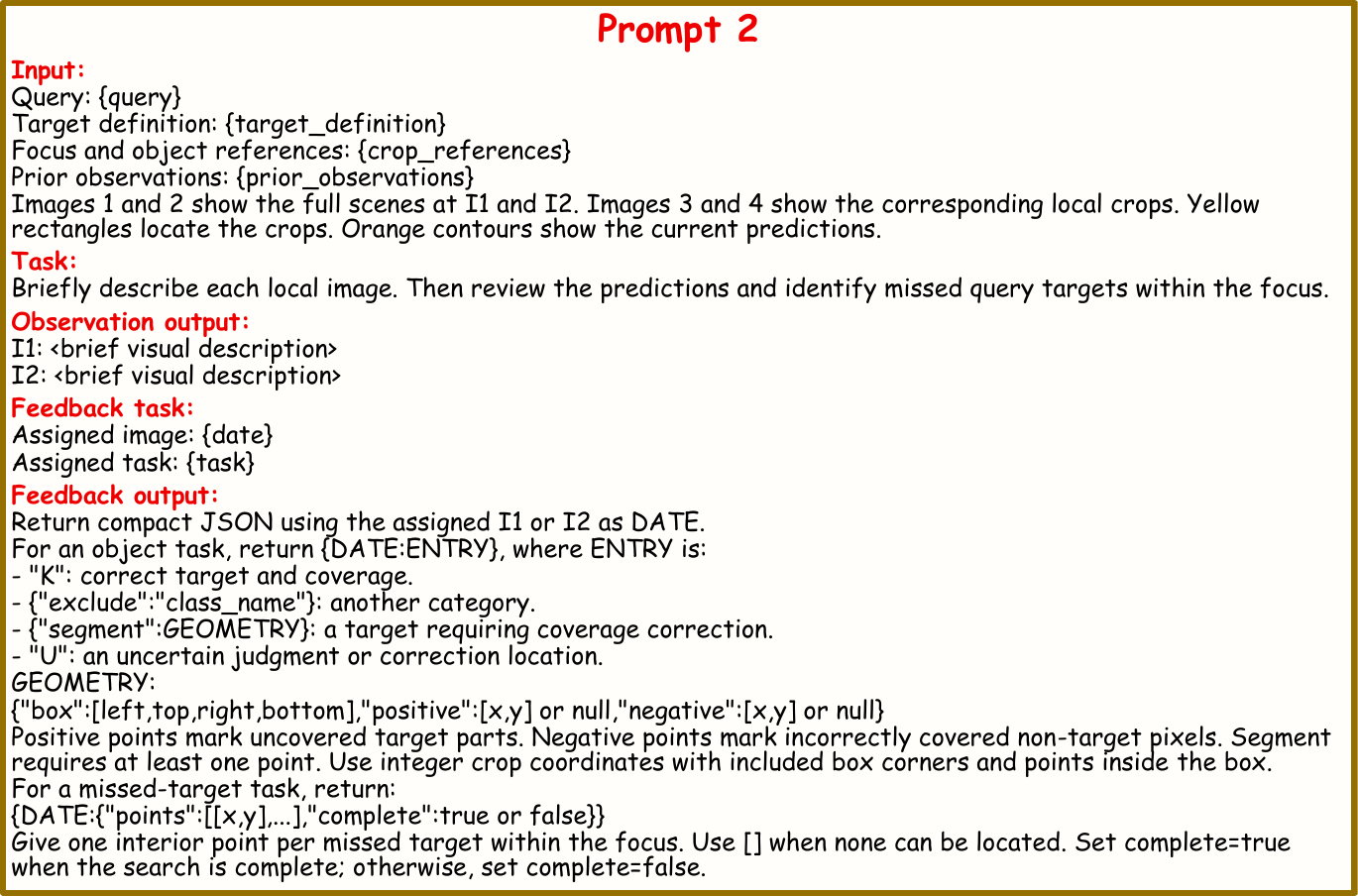}
    \caption{Prompt 2.}
    \label{fig:ablation_prompt2}
\end{figure*}

\noindent\textbf{Qwen2.5-VL prompts.}
Figure~\ref{fig:global_review_prompt} shows the global review prompt.
Figures~\ref{fig:local_observation_prompt}, \ref{fig:object_decision_prompt},
and~\ref{fig:missed_target_search_prompt} show the three prompts used
for local review. Figures~\ref{fig:ablation_prompt1} and~\ref{fig:ablation_prompt2}
show Prompt~1 and Prompt~2 used in the prompt ablation study.

\section{Additional Method Details}
\label{sec:supp_method}

This section provides additional details on review region construction,
feedback execution, and evidence reuse in LookBackOVCD.
Algorithm~\ref{alg:supp_inference} summarizes the inference procedure
of LookBackOVCD.
We follow the notation of the main paper.
The initial target responses are denoted by $R_t^{(0)}$, and their
displayed masks are
$M_t^{(0)}=\mathbf{1}[R_t^{(0)}>0.5]$,
where $t\in\{1,2\}$ indexes the two images.

\subsection{Review Region Construction}
\label{sec:supp_regions}

Connected regions in
$M_1^{(0)}\cup M_2^{(0)}$
provide the initial observation candidates.
Each candidate collects the corresponding predicted objects
from both images.
Nearby candidates are grouped to share an observation window,
providing surrounding context for reviewing individual objects.

We measure spatial proximity using the gaps between bounding boxes.
For candidates $i$ and $j$, let
$b_i=(x_i^{-},y_i^{-},x_i^{+},y_i^{+})$
and
$b_j=(x_j^{-},y_j^{-},x_j^{+},y_j^{+})$
denote their boxes.
Their distance is
\begin{equation}
\begin{aligned}
\delta_x(i,j)
&=
\max\{0,x_i^{-}-x_j^{+},x_j^{-}-x_i^{+}\},\\
\delta_y(i,j)
&=
\max\{0,y_i^{-}-y_j^{+},y_j^{-}-y_i^{+}\},\\
d(i,j)
&=
\sqrt{\delta_x(i,j)^2+\delta_y(i,j)^2}.
\end{aligned}
\label{eq:supp_box_distance}
\end{equation}
Overlapping or touching boxes therefore have zero distance.
A minimum spanning tree connects the candidates using these distances.
The tree edges are separated according to the maximum between-class
variance of their gap lengths, with the median longer side of the
candidate boxes providing a lower bound for the separation threshold.
Removing edges above this threshold produces spatial groups.

Let $H$ and $W$ denote the image height and width.
A group containing multiple candidates is divided when its bounding
box exceeds $W/2$ in width or $H/2$ in height.
The division follows the axis with the greatest spread of candidate
centers and splits the ordered candidates into two parts.
This process continues until the groups meet the local scale constraint.
A single large candidate is retained intact, allowing extended targets,
such as a large water body, to be observed as a whole.

Each group is expanded to include surrounding context.
The expansion depends on the member sizes and is bounded by the
image dimensions and the applicable local scale.
The resulting window contains the complete group and is applied
at the same location in both images.
Each existing object is assigned to one review group.
Neighboring windows can share contextual content while retaining
their respective object assignments.

\subsection{Review and Feedback Execution}
\label{sec:supp_feedback}

\noindent\textbf{Global review and routing.}
Qwen2.5-VL first examines both full images together with the initial
contours and candidate regions.
It assesses the target predictions and identifies regions requiring
local inspection.
It can also propose regions containing potentially overlooked targets,
extending review beyond the initial detections.

A region bypasses local review when its predictions are explicitly
resolved in both images.
For each image, the outlined objects must match the query with
adequate coverage and no suspected omission, or the region must
contain neither a queried target nor a predicted contour.
Confirmed global decisions proceed directly to feedback execution.
The remaining regions enter local review.

\medskip
\noindent\textbf{Local review.}
For each window $w_r$, the visual input contains the full images
and paired local crops.
The full images provide scene context and indicate the crop location,
while the local views expose target appearance and mask boundaries.
The query, assigned object references, and retrieved evidence
accompany these views.

The model first produces a brief observation for each image,
examining object attributes, surrounding context, and mask coverage.
It then generates a separate decision for each assigned object
and searches for queried targets without predicted contours.
These decisions share the cached visual context.
Objects in the same crop can therefore receive different actions
without separate image encodings.

\begin{algorithm}[t]
\caption{Inference procedure of LookBackOVCD}
\label{alg:supp_inference}
\small
\begin{algorithmic}[1]
\Require Image pair $I_1,I_2$, text query $q$, change threshold $\tau$
\Ensure Change mask $\widehat{Y}_q$

\State Obtain feature maps $F_t$ and instance predictions $S_t^q$
using frozen SAM~3
\State Construct initial responses $R_t^{(0)}$, displayed masks
$M_t^{(0)}$, and temporal comparison fields $G_t^{(0)}$
\State $R^{(0)}\gets(R_1^{(0)},R_2^{(0)})$,
$G^{(0)}\gets(G_1^{(0)},G_2^{(0)})$
\State $\mathcal{W}\gets
\Call{ConstructReviewRegions}{M_1^{(0)}\cup M_2^{(0)}}$
\State $\begin{aligned}[t]
&(\mathcal{O}_g,\mathcal{D}_g,\mathcal{W}_{\mathrm{loc}})\gets\\
&\quad\Call{GlobalReviewAndRouting}{I_1,I_2,q,R^{(0)},\mathcal{W}}
\end{aligned}$
\State $\mathcal{B}\gets\Call{StoreGlobalObservations}{\mathcal{O}_g}$
\State Map the spatial prompts in $\mathcal{D}_g$ to the original
image coordinates
\State $\begin{aligned}[t]
&\mathcal{P}_g\gets\Call{GenerateFeedbackProposals}{}\\
&\quad\bigl(I_1,I_2,R^{(0)},\mathcal{D}_g,\text{SAM~3}\bigr)
\end{aligned}$
\State $\begin{aligned}[t]
&(R^{(1)},G^{(1)})\gets\\
&\quad\Call{MergeAndUpdate}{R^{(0)},G^{(0)},F_1,F_2,\mathcal{P}_g}
\end{aligned}$
\State $(\widehat{R},\widehat{G})\gets(R^{(1)},G^{(1)})$

\If{$\mathcal{W}_{\mathrm{loc}}\ne\varnothing$}
    \State $\mathcal{D}_{\mathrm{loc}}\gets\varnothing$
    \For{each batch $\mathcal{W}_b$ of $\mathcal{W}_{\mathrm{loc}}$}
        \ForAll{$w_r\in\mathcal{W}_b$}
            \State $\mathcal{E}^{(r)}\gets
            \Call{RetrieveEvidence}{\mathcal{B},w_r,R^{(1)}}$
            \State $\mathcal{V}^{(r)}\gets
            \Call{PrepareViews}{I_1,I_2,w_r,R^{(1)}}$
            \State $\begin{aligned}[t]
            &(\mathcal{O}^{(r)},\mathcal{D}^{(r)})\gets\\
            &\quad\Call{LocalReview}{\mathcal{V}^{(r)},q,\mathcal{E}^{(r)}}
            \end{aligned}$
        \EndFor
        \State Add the batch observations and review states to $\mathcal{B}$
        \State $\mathcal{D}_{\mathrm{loc}}\gets
        \mathcal{D}_{\mathrm{loc}}\cup
        \bigcup_{w_r\in\mathcal{W}_b}\mathcal{D}^{(r)}$
    \EndFor
    \State Map the spatial prompts in $\mathcal{D}_{\mathrm{loc}}$
    to the original image coordinates
    \State $\begin{aligned}[t]
    &\mathcal{P}_{\mathrm{loc}}\gets\Call{GenerateFeedbackProposals}{}\\
    &\quad\bigl(I_1,I_2,R^{(1)},\mathcal{D}_{\mathrm{loc}},\text{SAM~3}\bigr)
    \end{aligned}$
    \State $\begin{aligned}[t]
    &(R^{(2)},G^{(2)})\gets\\
    &\quad\Call{MergeAndUpdate}{R^{(1)},G^{(1)},F_1,F_2,\mathcal{P}_{\mathrm{loc}}}
    \end{aligned}$
    \State $(\widehat{R},\widehat{G})\gets(R^{(2)},G^{(2)})$
\EndIf

\State $\widehat{Y}_q\gets
\Call{TemporalComparison}{\widehat{R},\widehat{G},\tau}$
\State \Return $\widehat{Y}_q$
\end{algorithmic}
\end{algorithm}

\medskip
\noindent\textbf{Spatial prompts.}
Feedback associates each decision with its image and the object
requiring correction.
For an existing object, \textit{Keep} and \textit{Exclude} reuse
its box, while \textit{Segment} supplies a correction box and
positive or negative points.
Positive points indicate target foreground, and negative points
identify non-target pixels included in the prediction.

Spatial prompts are mapped from the displayed view to the original
image coordinates.
For a crop
$w_r=(x_r^0,y_r^0,x_r^1,y_r^1)$
displayed at size
$(\widetilde{W}_{t,r},\widetilde{H}_{t,r})$,
a point $(u,v)$ is mapped by
\begin{equation}
\begin{aligned}
x
&=
x_r^0+
\frac{x_r^1-x_r^0}{\widetilde{W}_{t,r}}u,\\
y
&=
y_r^0+
\frac{y_r^1-y_r^0}{\widetilde{H}_{t,r}}v.
\end{aligned}
\label{eq:supp_coordinate_mapping}
\end{equation}
Boxes follow the same transformation.
Their endpoints are converted to the image indexing convention,
with outward rounding to preserve the specified region.

\medskip
\noindent\textbf{Response updates.}
SAM~3 converts the spatial prompts into mask probabilities.
For \textit{Segment}, the prompted probability map updates the
target response within the object box.
For \textit{Exclude}, a box prompt generates a foreground mask.
Its union with the current predicted foreground, restricted to the
box, defines the region whose response is set to zero.

For \textit{Keep}, the box-prompted mask provides a second boundary
estimate.
When its IoU with the current mask exceeds $0.5$, their probabilities
are averaged along the immediate boundary of the current contour.
The boundary band is determined by foreground--background
transitions between horizontally or vertically adjacent pixels.

For a potentially missed target, Qwen provides an interior positive
point.
SAM~3 directly generates candidate masks from this point, and the
mask with the highest predicted quality is selected among those
containing it.
Its newly recovered foreground is added to the target response.

Feedback proposals are constructed from the same current predictions
and applied within their corresponding correction regions.
Compatible updates are merged, while pixels receiving conflicting
updates retain their previous values.
Uncertain judgments are recorded with the corresponding predictions
and review states.

Accepted corrections update the target responses.
When an object mask changes, its representations are recomputed
from the retained feature maps $F_1$ and $F_2$ over the updated support.
The associated temporal comparison fields are updated accordingly.
The final change map is obtained from the corrected responses
and temporal feature differences using the comparison rule in
the main paper.

\subsection{Evidence Pool Management}
\label{sec:supp_evidence_management}

A shared evidence pool is maintained for each image pair and query.
Global and local observations are stored with their image indices,
region coordinates, and review states.
Regional records also retain the associated object references.
The pool is updated as review proceeds, making earlier observations
available to later regions within the same image pair.

Scene observations provide global context.
For a local crop, regional evidence is retrieved according to
spatial overlap and its correspondence with the current prediction
state.
A response signature associates each regional record with the
predictions observed when it was created.
Object references are filtered by intersection with the current
crop, and their coordinates are expressed in the original image
coordinate system.

The retrieved evidence provides earlier observations of object
appearance, contextual relationships, and mask coverage.
Qwen combines these observations with the current visual input
to assess the assigned objects and generate feedback.
Regions processed in the same batch use the available pool state,
and their observations are added for subsequent batches.
SAM~3 uses the decisions from local review to update the target
responses before temporal comparison.

\section{More Diagnostic Experiments}
\label{sec:supp_experiments}

\subsection{Error Analysis Before and After Feedback}
\label{sec:supp_feedback_errors}

\begin{figure*}[t]
    \centering
    \includegraphics[width=\textwidth]{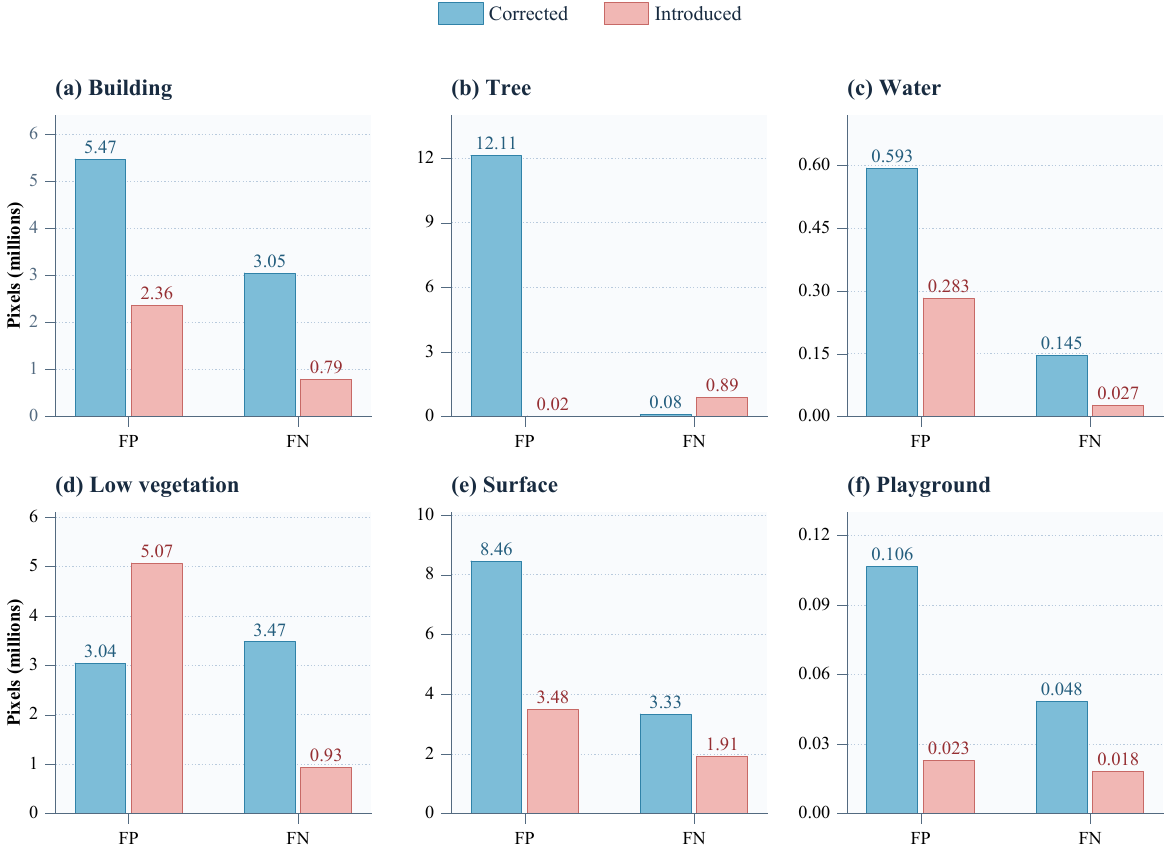}
    \caption{Pixel errors before and after feedback on SECOND.
    Blue and red bars show corrected and introduced errors,
    respectively. Counts are in millions of pixels, with a separate
    vertical scale for each class.}
    \label{fig:supp_feedback_errors}
\end{figure*}

Figure~\ref{fig:supp_feedback_errors} compares the SAM~3 baseline
and LookBackOVCD on the 593 test pairs of SECOND. For each class,
we compare both change masks with the ground truth and count the
pixels where false positives and false negatives are corrected or
introduced.

Both error counts decrease for building, water, surface, and
playground. For the tree class, false positives decrease by
12.09 million pixels, while false negatives increase by
0.81 million. Low vegetation shows the opposite pattern, with
2.54 million fewer false negatives and 2.03 million more false
positives. Where tree crowns touch or vegetation forms continuous
patches, boxes and points may provide limited guidance on where
a mask should be extended or trimmed.

\subsection{Effects of Global and Local Review}
\label{sec:supp_review_ablation}
Table~\ref{tab:supp_review_ablation} compares global review,
local review, and their combination on SECOND and SC-SCD.
The global-only setting reviews the two full images, while the
local-only setting uses only paired crops of the initial candidate
regions. Local review alone achieves higher scores than global
review alone on both datasets. The combined setting achieves the
highest scores, improving $\mathrm{mIoU}_{C}$ and $\mathrm{mF1}_{C}$
over local review alone by 2.3 and 2.5 percentage points on SECOND
and by 1.5 and 1.9 percentage points on SC-SCD. These results
suggest that combining scene context from the full images with
object details from the local crops provides more complete evidence
for reviewing target predictions and improves change detection accuracy.

\begin{table}[t]
\centering
\caption{Effects of global and local review on SECOND and SC-SCD.
All values are percentages.}
\label{tab:supp_review_ablation}

\small
\setlength{\tabcolsep}{3pt}
\renewcommand{\arraystretch}{1.15}
\begin{tabular*}{\columnwidth}{@{\extracolsep{\fill}}lcccc@{}}
\toprule
\multirow{2}{*}{\textbf{Review setting}}
& \multicolumn{2}{c}{\textbf{SECOND}}
& \multicolumn{2}{c}{\textbf{SC-SCD}} \\
\cmidrule(lr){2-3}\cmidrule(lr){4-5}
& $\mathrm{mIoU}_{C}$ & $\mathrm{mF1}_{C}$
& $\mathrm{mIoU}_{C}$ & $\mathrm{mF1}_{C}$ \\
\midrule
Global only                 & 30.3 & 45.9 & 24.6 & 38.1 \\
Local only                  & 32.6 & 48.4 & 25.9 & 39.7 \\
\textbf{Global + local}     & 34.9 & 50.9 & 27.4 & 41.6 \\
\bottomrule
\end{tabular*}
\end{table}

\section{More Qualitative Results}
\label{sec:supp_qualitative}
\begin{figure*}[tp]
    \centering
    \includegraphics[width=\textwidth]{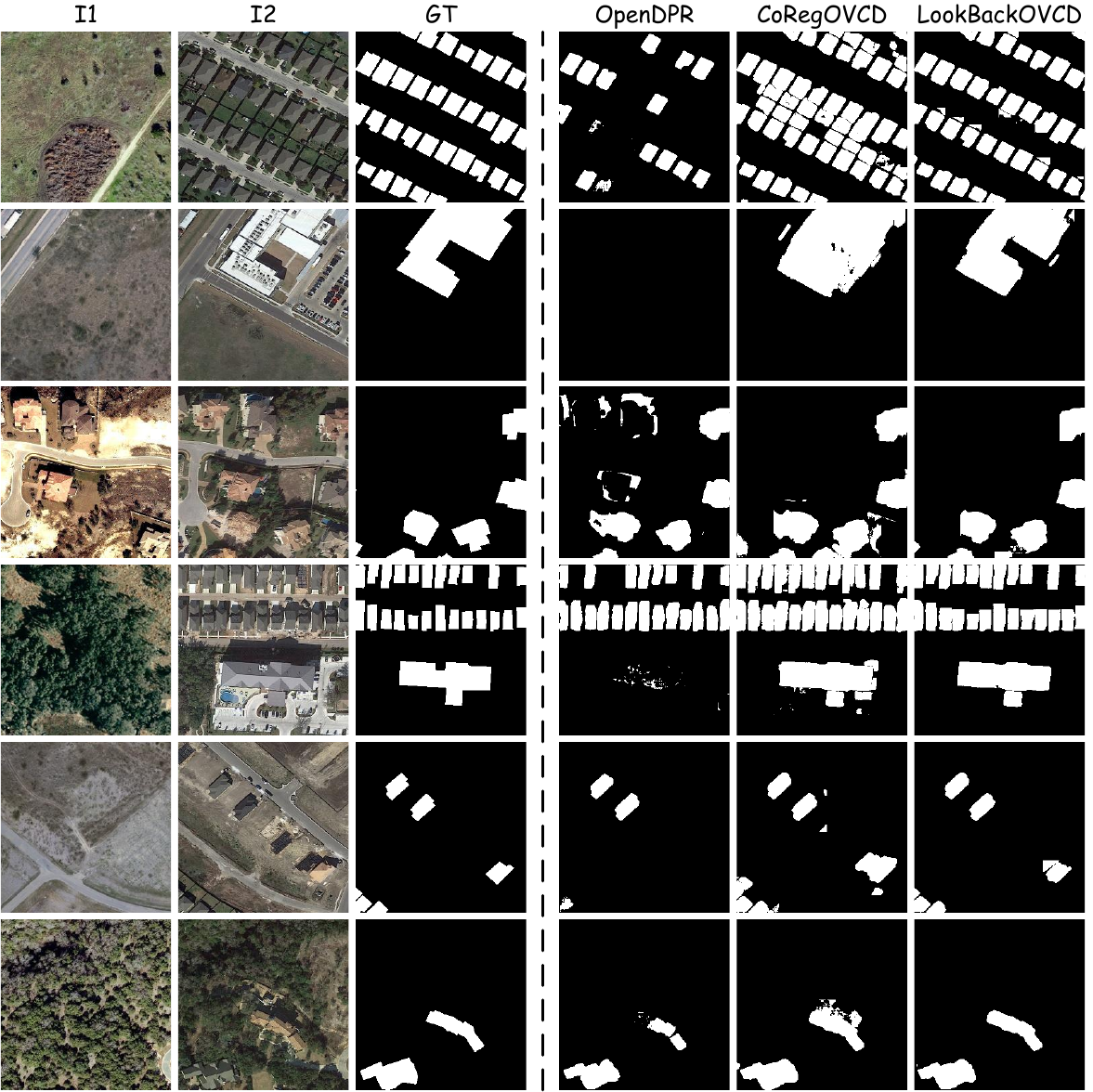}
    \caption{Qualitative comparison on LEVIR-CD.}
    \label{fig:supp_qual_levir}
\end{figure*}

\begin{figure*}[tp]
    \centering
    \includegraphics[width=\textwidth]{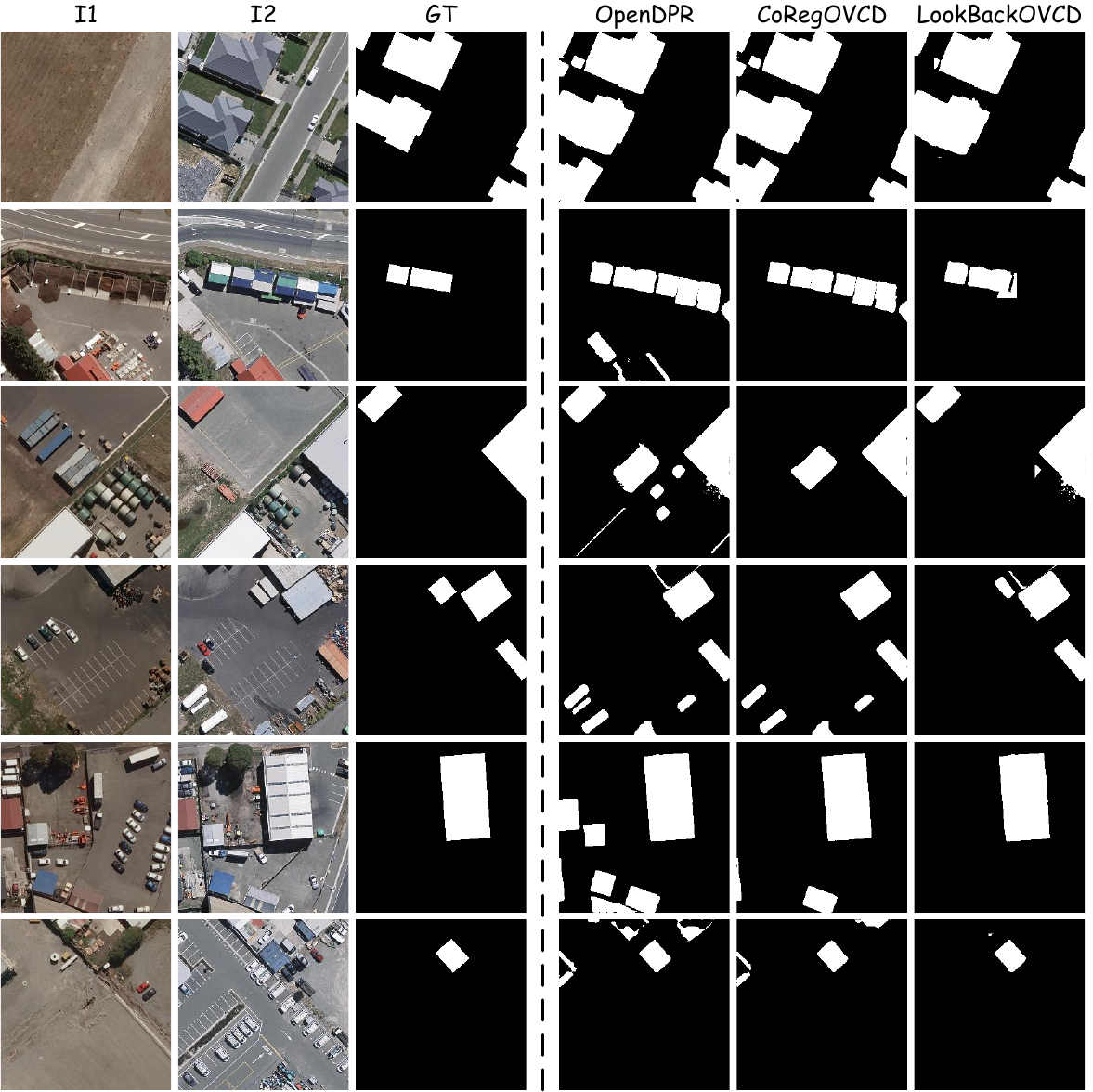}
    \caption{Qualitative comparison on WHU-CD.}
    \label{fig:supp_qual_whu}
\end{figure*}

\begin{figure*}[tp]
    \centering
    \includegraphics[width=\textwidth]{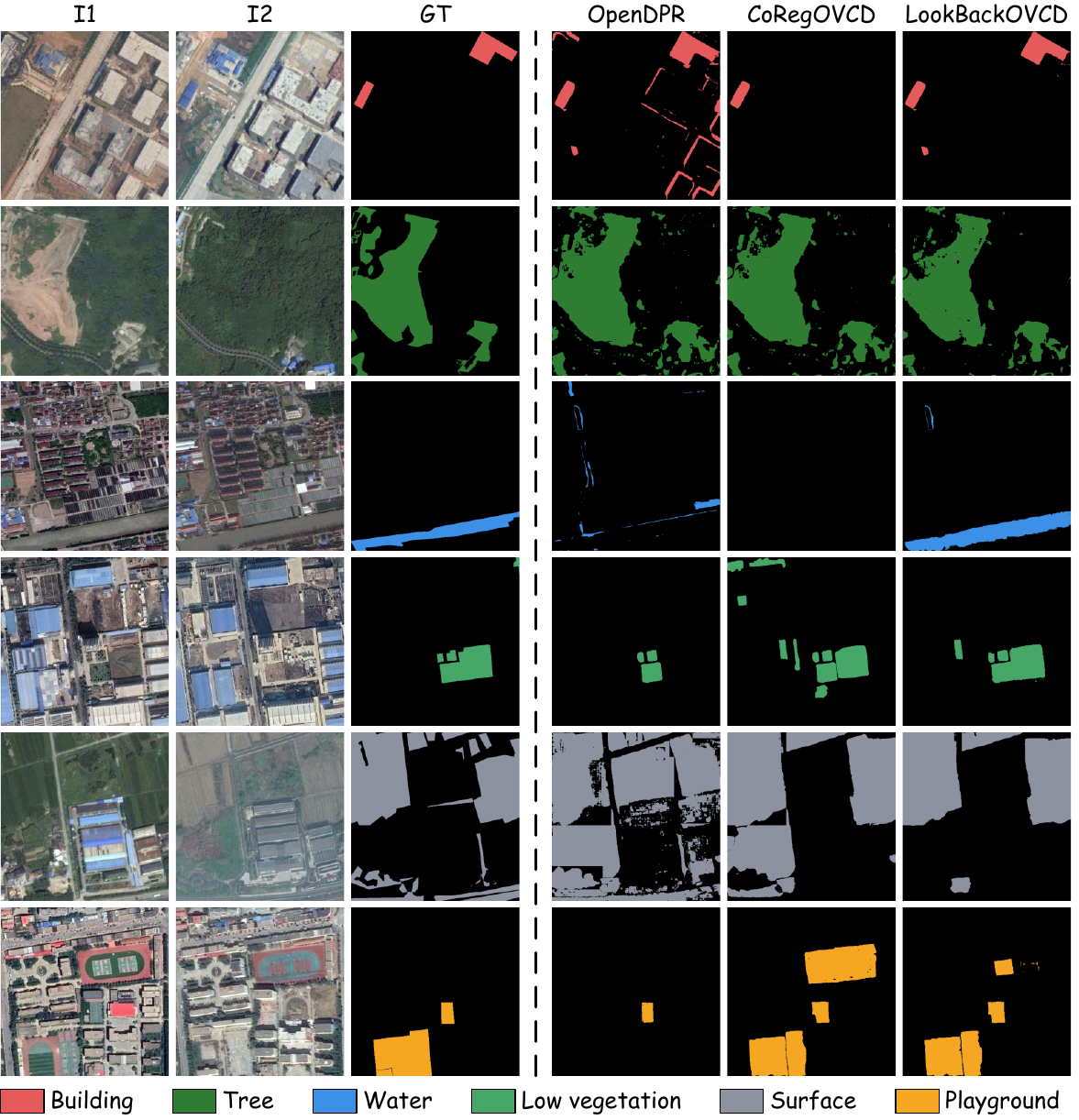}
    \caption{Qualitative comparison on SECOND.}
    \label{fig:supp_qual_second}
\end{figure*}

\begin{figure*}[tp]
    \centering
    \includegraphics[width=\textwidth]{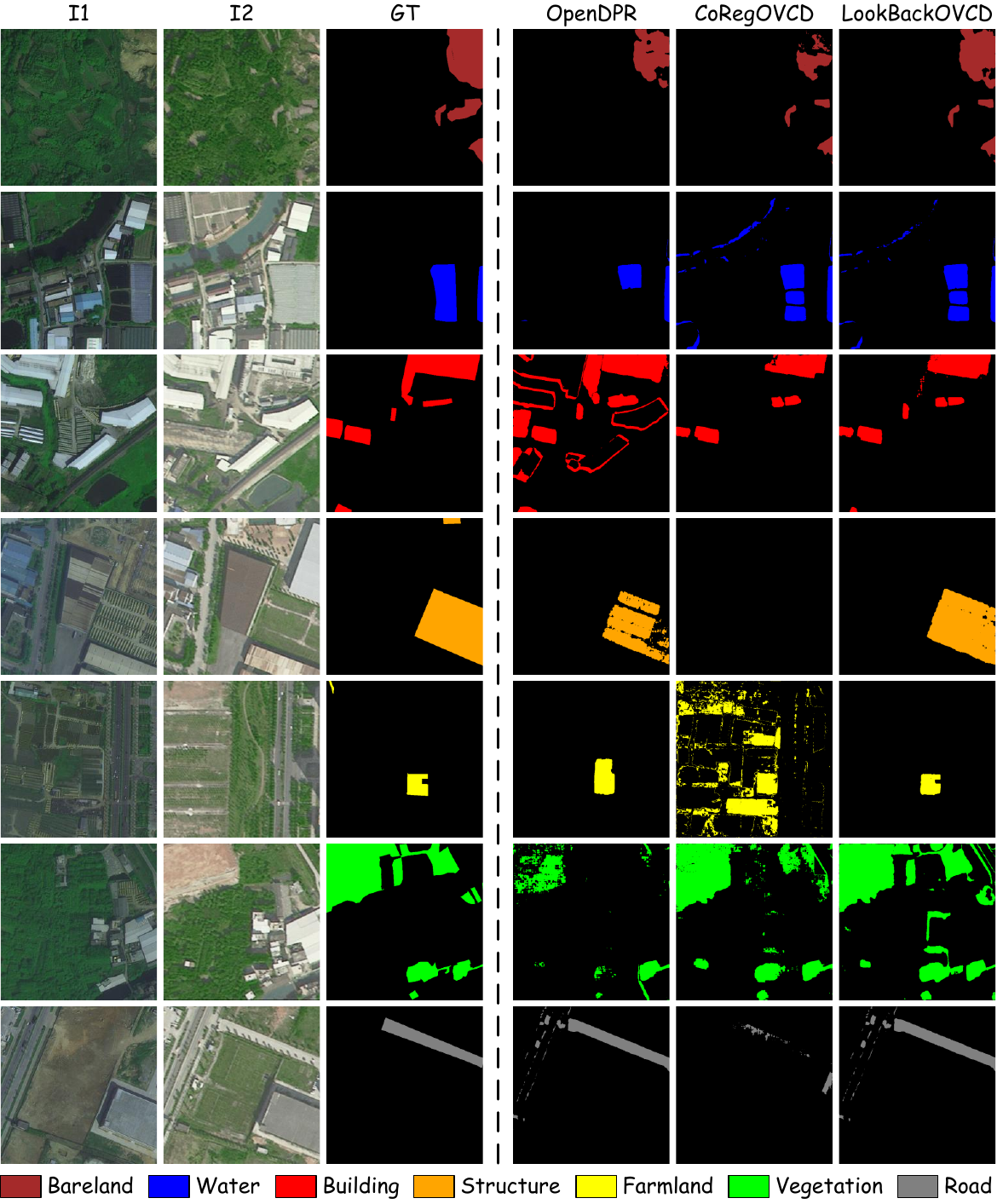}
    \caption{Qualitative comparison on SC-SCD.}
    \label{fig:supp_qual_scd}
\end{figure*}

Figures~\ref{fig:supp_qual_levir}, \ref{fig:supp_qual_whu},
\ref{fig:supp_qual_second}, and \ref{fig:supp_qual_scd} compare
LookBackOVCD with OpenDPR and CoRegOVCD on LEVIR-CD, WHU-CD,
SECOND, and SC-SCD. Each row shows the two input images, the
ground truth, and the predicted change masks. In the LEVIR-CD
examples, LookBackOVCD recovers missed buildings and better
preserves the gaps between adjacent roofs. The examples on
SECOND and SC-SCD show results for several land-cover categories.

\end{document}